\documentclass[journal,twoside]{IEEEtran}

\IEEEoverridecommandlockouts    %

\usepackage{times}              %
\usepackage{amsmath}            %
\usepackage{amssymb}            %
\usepackage{bm}                 %

\usepackage{array}
\usepackage{epsfig}
\usepackage{etoolbox}
\usepackage{graphics}
\usepackage{makecell}
\usepackage{microtype}
\usepackage{stfloats}
\usepackage{epstopdf} %

\usepackage[noadjust]{cite}     %
\ifCLASSOPTIONcompsoc
    \usepackage[caption=false, font=normalsize, labelfont=sf, textfont=sf]{subfig}
\else
    \usepackage[caption=false, font=footnotesize]{subfig}
\fi

\makeatletter
\let\NAT@parse\undefined
\makeatother
\usepackage[bookmarks=true,hidelinks]{hyperref}

\usepackage[noabbrev]{cleveref}      %
\Crefname{section}{Section}{Section}
\Crefname{subsection}{Section}{Section}
\Crefname{subsubsection}{Section}{Section}
\Crefname{figure}{Figure}{Figure}

\usepackage{paralist}           %

\usepackage[binary-units=true]{siunitx}

\newcommand{\vx}{\bm{x}}

\newcommand{\vu}{\bm{u}}
\newcommand{\vf}{\bm{f}}

\newcommand{\argmin}{\operatornamewithlimits{arg\ min}}
\newcommand{\orcid}[1]{\href{https://orcid.org/#1}{\includegraphics[width=0.6em]{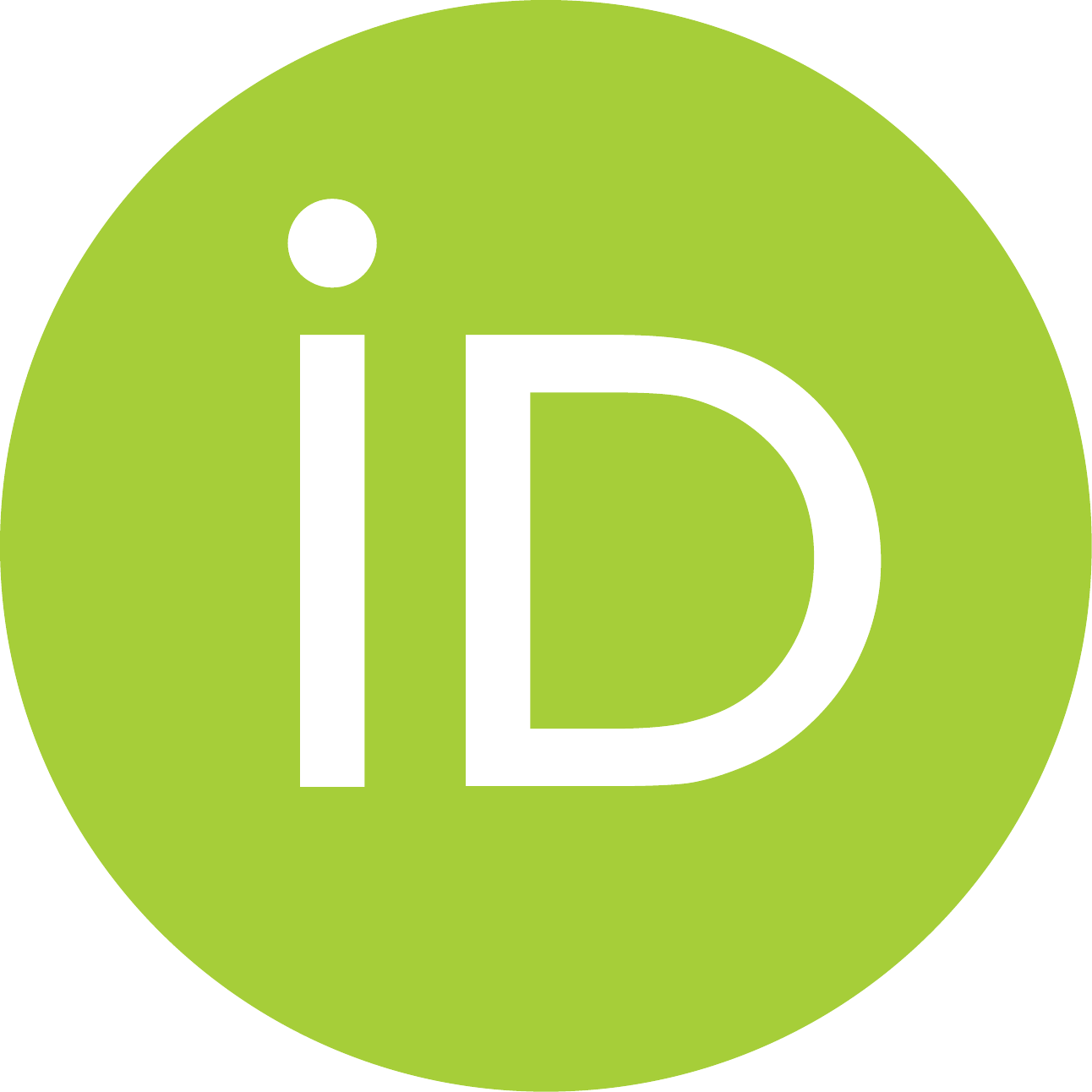}}}

\usepackage[latin1]{inputenc}   %

\usepackage{graphicx}
\graphicspath{{figures/}}
\usepackage[export]{adjustbox} %

\usepackage{algorithm}
\usepackage{algorithmic}
\makeatletter
\newcommand\fs@norules{\def\@fs@cfont{\bfseries}\let\@fs@capt\floatc@ruled
  \def\@fs@pre{}%
  \def\@fs@post{}%
  \def\@fs@mid{\kern3pt}%
  \let\@fs@iftopcapt\iftrue}
\makeatother
\floatstyle{norules}
\restylefloat{algorithm}

\usepackage[style=long,nolist]{glossaries}
\glsdisablehyper
\newacronym{c-space}{$\mathcal{C}$-space}{configuration space}
\newacronym{com}{CoM}{Center-of-Mass}
\newacronym{dof}{DoF}{degrees of freedom}
\newacronym{eom}{EoM}{equation of motion}
\newacronym{fk}{FK}{forward kinematics}
\newacronym{giw}{GIW}{Gravito-Inertial Wrench}
\newacronym{ik}{IK}{inverse kinematics}

\newacronym{hmm}{HMM}{Hidden Markov Model}

\newacronym{lq}{LQ}{Linear-Quadratic}
\newacronym{lp}{LP}{linear program}
\newacronym{lping}{LP}{Linear Programming}
\newacronym{nlp}{NLP}{Non-linear Programming}
\newacronym{qp}{QP}{Quadratic Programming}
\newacronym{sdp}{SDP}{Semidefinite Programming}
\newacronym{sqp}{SQP}{Sequential Quadratic Programming}
\newacronym{miqcqp}{MIQCQP}{Mixed-Integer Quadratically Constrained Quadratic Programme}
\newacronym{sip}{SIP}{Semi-Infinite Programming}

\newacronym{rrt}{RRT}{Rapidly-Exploring Random Tree}
\newacronym{ddp}{DDP}{Differential Dynamic Programming}
\newacronym{fddp}{FDDP}{Feasibility-driven Differential Dynamic Programming}
\newacronym{boxddp}{BoxDDP}{Control-limited Differential Dynamic Programming}
\newacronym{boxfddp}{BoxFDDP}{Control-limited Feasibility-driven Differential Dynamic Programming}
\newacronym{sbp}{SBP}{Sampling-based Planning}
\newacronym{knn}{KNN}{$k$-Nearest Neighbor}
\newacronym{pca}{PCA}{Principal Component Analysis}
\newacronym{mse}{MSE}{Mean Squared Error}
\newacronym{mae}{MAE}{Mean Absolute Error}
\newacronym{lwpr}{LWPR}{Locally Weighted Projection Regression}
\newacronym{gpr}{GPR}{Gaussian Process Regression}
\newacronym{mlp}{MLP}{Multi-Layer Perceptron}

\newacronym{mpc}{MPC}{Model-Predictive Control}
\newacronym{pbd}{PbD}{Programming by Demonstration}
\newacronym{lfd}{LfD}{Learning from Demonstration}
\newacronym{ioc}{IOC}{Inverse Optimal Control}
\newacronym{irl}{IRL}{Inverse Reinforcement Learning}
\newacronym{oc}{OC}{Optimal Control}
\newacronym{rl}{RL}{Reinforcement Learning}

\newacronym{moe}{MoE}{Mixture-of-Experts}
\newacronym{poe}{PoE}{Product-of-Experts}
\newacronym{gmm}{GMM}{Gaussian Mixture Model}

\newacronym{agv}{AGV}{Autonomous Ground Vehicle}
\newacronym{ros}{ROS}{Robot Operating System}
\newacronym{exotica}{EXOTica}{Extensible Optimization Toolset}

\usepackage[english]{babel}

\hypersetup{
    pdfauthor   = {Wolfgang Merkt and Vladimir Ivan},%
    pdftitle    = {Memory Clustering using Persistent Homology for Multimodality- and Discontinuity-Sensitive Learning of Optimal Control Warm-starts},%
    pdfsubject  = {Robotics},%
    pdfkeywords = {Robots, Optimal Control, Persistent Homology, Clustering, Mixture-of-Experts, Memory-of-Motion}%
}

\title{
Memory Clustering using Persistent Homology for Multimodality- and Discontinuity-Sensitive Learning of Optimal Control Warm-starts
}

\author{
    Wolfgang Merkt$^{*}$\,\orcid{0000-0003-3235-4906}\quad
    Vladimir Ivan$^{*}$\,\orcid{https://orcid.org/0000-0002-6610-385X}\quad 
    Traiko Dinev\,\orcid{0000-0002-4099-2775}\quad
    Ioannis Havoutis\,\orcid{0000-0002-4371-4623}\quad
    Sethu Vijayakumar\,\orcid{0000-0003-0649-7241}%
\thanks{${}^{*}$ The first two authors contributed equally to this work.}
\thanks{%
This research was supported by (1) the European Commission under the Horizon 2020 project Memory of Motion (MEMMO, project ID: 780684), (2) the Engineering and Physical Sciences Research Council (EPSRC) UK RAI Hub for Offshore Robotics for Certification of Assets (ORCA, grant reference EP/R026173/1), and (3) the EPSRC Centre for Doctoral Training in Robotics and Autonomous Systems (EPSRC, grant reference EP/L016834/1).
\textit{(Corresponding author: Wolfgang Merkt.)}
}%
\thanks{Wolfgang Merkt and Ioannis Havoutis are with the Oxford Robotics Institute, University of Oxford, Oxford, OX2 6NN, U.K. (e-mail: wolfgang@robots.ox.ac.uk; ioannis@robots.ox.ac.uk).}%
\thanks{Vladimir Ivan, Traiko Dinev, and Sethu Vijayakumar are with the School of Informatics, The University of Edinburgh, Edinburgh EH8 9AB, U.K. (e-mail: v.ivan@ed.ac.uk; traiko.dinev@ed.ac.uk; sethu.vijayakumar@ed.ac.uk).}%
}

\begin{document}
\bstctlcite{IEEEexample:BSTcontrol}

\maketitle
\thispagestyle{empty}
\pagestyle{empty}

\begin{abstract}
    Shooting methods are an efficient approach to solving nonlinear optimal control problems.
As they use local optimization, they exhibit favorable convergence when initialized with a good warm-start but may not converge at all if provided with a poor initial guess.
Recent work has focused on providing an initial guess from a learned model trained on samples generated during an offline exploration of the problem space.
However, in practice the solutions contain discontinuities introduced by system dynamics or the environment. Additionally, in many cases multiple equally suitable, i.e., multi-modal, solutions exist to solve a problem.
Classic learning approaches smooth across the boundary of these discontinuities and thus generalize poorly.
In this work, we apply tools from algebraic topology to extract information on the underlying structure of the solution space. In particular, we introduce a method based on persistent homology to automatically cluster the dataset of precomputed solutions to obtain different candidate initial guesses. 
We then train a Mixture-of-Experts within each cluster to predict state and control trajectories to warm-start the optimal control solver and provide a comparison with modality-agnostic learning.
We demonstrate our method on a cart-pole toy problem and a quadrotor avoiding obstacles, and show that clustering samples based on inherent structure improves the warm-start quality. 

\end{abstract}

\section{Introduction}
\IEEEPARstart{O}{ptimal}
control can be used to generate highly dynamic motions and behaviors by specifying objectives composed of desirable characteristics with dynamics taken as constraints.
In particular, it allows composing complex maneuvers by working over a long time horizon. In contrast, instantaneous control methods such as inverse dynamics are unable to solve such problems.
Examples of this include swing-up of an under-actuated cartpole or jumps and front-flips on legged platforms \cite{mastalli2020crocoddyl}.
However, optimal control methods can take longer to converge if not warm-started and can easily get stuck in a local minimum if provided with a poor initial guess.
Previous work has focused on exploration of the parameterized optimal control problem in an offline computation process to gather experience to initialize from during runtime \cite{atkeson2003nonparametric,stolle2006policies,tassa2008receding,jetchev2013fast,dey2013contextual,mansard2017irepa,merkt2018leveraging,tang2019discontinuity,lembono2020memory}.

\begin{figure}[t]
    \centering
    \includegraphics[width=\linewidth]{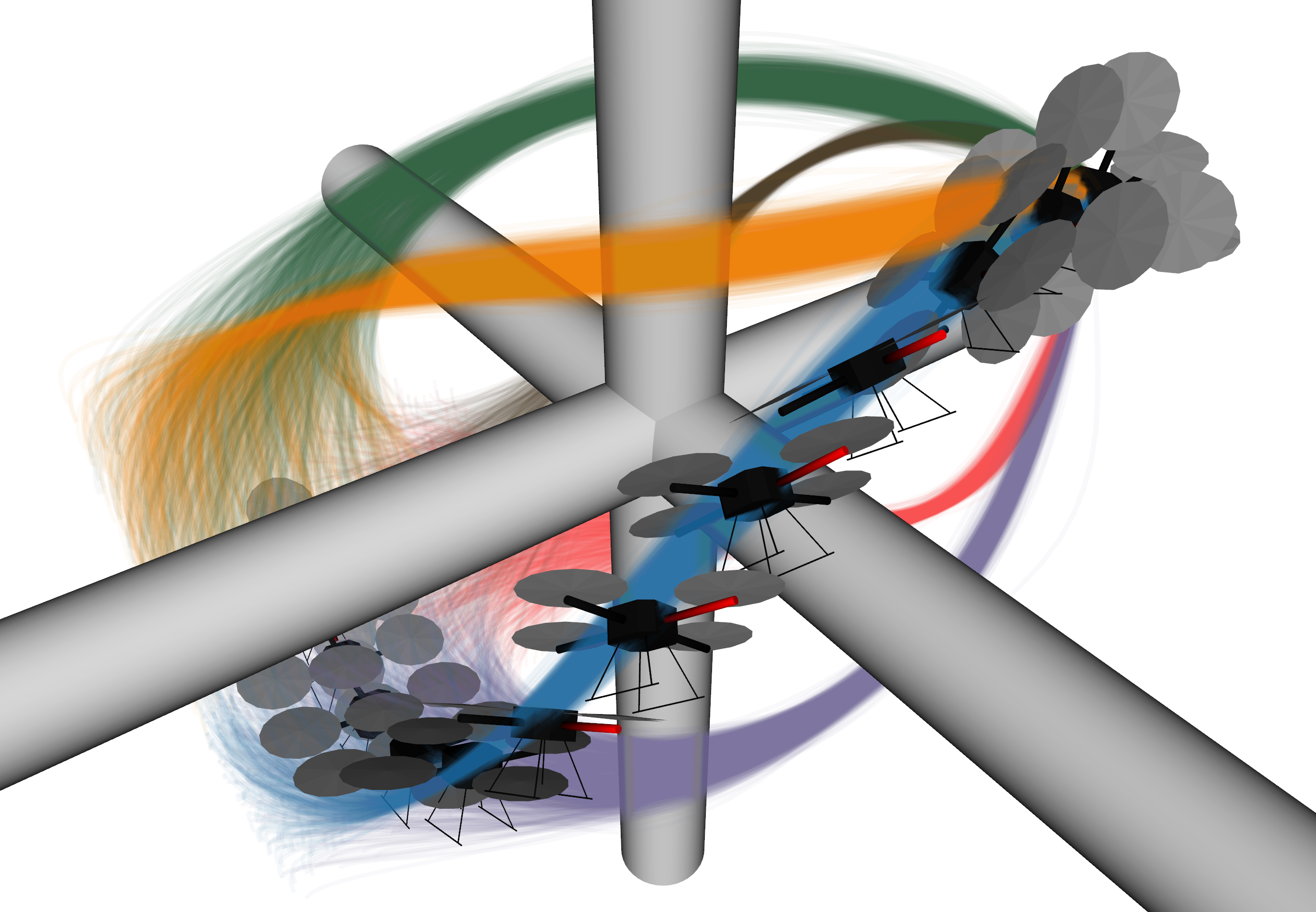}
    \caption[Illustration of discontinuity and multi-modality]{Illustration of discontinuity and multi-modality: 
    Quadcopter flying in a complex environment: The different classes of trajectories from start to goal cannot be continuously deformed into each one another.
    This violates the continuous map assumption between problem parametrization and solution output $f : X \to Y$ which is core to function approximation. The number of clusters and class memberships are automatically extracted using our method.}
    \label{fig:discontinuity_multimodality}
\end{figure}

When considering optimal control problems, one can observe two cases that make it challenging to directly use learning as a way of compressing and generalizing across optimal solution samples:
First, \emph{discontinuity}, where similar problems (in parameter space) yield vastly different optimal solutions. Examples of this can be seen in a) the phase space plot of optimal solutions for a pendulum or cartpole swing-up task and b) with discontinuities in solution paths introduced by environment obstacles.
Second, \emph{multi-modality}, where multiple equally optimal solutions to a problem exist. A prominent example is the ability to traverse around an obstacle in multiple ways as seen in \Cref{fig:discontinuity_multimodality}.%
\footnote{We note that in some cases one mode may be less optimal than another, however, as most work relies on locally optimal samples generated using a stochastic precomputation process, this may not be known a priori. Additionally, storing multiple modes can increase warm-start success and robustness as one mode may, for instance, be obstructed by an obstacle.}

Both discontinuity and multi-modality can greatly impact the quality of prediction obtained using function approximation as regressors smooth across the boundaries between clusters or modalities.
This is expected as efficient function approximation methods such as \gls{lwpr} \cite{vijayakumar2000locally} and \gls{gpr} \cite{williams1996gaussian} assume uni-modal distributions and continuity.
One workaround to avoid multi-modality is to bias the sampling/exploration stage to include/enforce only one modality (e.g., \cite{mansard2017irepa}).
Alternatively, we can consider machine learning models that can handle multi-modality and discontinuity directly \cite{ToussaintDiscontinuities}.
These are often a combination of multiple local models that each model one continuous cluster of data well.
Clustering time-series data and learning trajectories has been widely explored under the paradigm of \gls{pbd} to extract a set of alternative, feasible solutions from a dataset of demonstration trajectories.
For input partitioning, hierarchical clustering has previously been applied using simple distance-based geometric approaches \cite{aleotti2006ras} and \glspl{hmm} \cite{oates1999clustering,ogawara2002modeling}. The former required heuristics to be set while the latter was reported by \cite{aleotti2006ras} to be unstable.
Previous work circumventing this issue used trajectory libraries queried with nearest neighbor \cite{atkeson2003nonparametric, stolle2006policies, stolle2007transfer, tassa2008receding, merkt2018leveraging}, a hyper-local model that returns the closest neighbor without any interpolation. Thus, the returned solution is valid (i.e., does not violate any constraints for a similar problem), but does not generalize between samples (and also does not compress the original dataset).
In order to improve success rates for handling discontinuous and multi-modal distributions directly, \gls{moe} \cite{jacobs1991mixtureofexperts} and \gls{poe} \cite{hinton1999productsofexperts} systems have been proposed.
In \gls{moe}, a gating function or network determines which expert will provide the best output. The rationale is that a regressor trained exclusively on a continuous subset of the data will do well when queried within that subset but poorly outside.
Traditionally, \gls{moe} systems first partition the data based on the similarity of the input-output mapping and then train experts individually on subsets of data.
If a clustering is not known, joint training using a loss function that encourages both specialization and cooperation between the local experts, and thus automatically assigns samples to classes, can be employed \cite{jacobs1991mixtureofexperts}.
\gls{poe} methods, on the other hand, combine the output of multiple probabilistic models to form a prediction.
An example is the \glspl{gmm} which can capture multi-modality directly, however, they require information on the number of classes present or the tuning of hyper-parameters. \cite{pignat2019bayesiangmm}, for instance, use a Dirichlet process which can possibly represent an infinite number of clusters, while \cite{bishop2006prml} applies a Dirichlet distribution prior with a fixed number of clusters.

The authors of \cite{tang2019discontinuity} used a \gls{moe} approach for discontinuity-sensitive learning of initialization seeds.
They applied k-means clustering informed by expert knowledge on the input-output relationship such as periodicity of angles or Lagrange multipliers of constraints. They then learned a \gls{moe} system and applied it on a pendulum, 2D car, and a quadrotor with a single spherical obstacle. %

We argue that while these approaches may work well on small problems and with datasets where a system designer's intuition is readily available, it is challenging to extract heuristics for labeling data (or to extract the number of clusters) on higher dimensional tasks.
Furthermore, due to the stochastic nature of the exploration stage, the dataset may not include samples of all topologically distinct classes or modalities.

The authors of \cite{pokorny2016} took a different approach to cluster trajectories. They used filtrations of simplicial complexes and persistent homology for modeling trajectories in configuration spaces. They then used the persistency of simplicial complexes to classify trajectories with fixed start and end points. This method looks at the changes in topology across different scales and identifies at which scale topological features (connected components and holes) appear and disappear. The theory behind this approach has been studied in \cite{edelsbrunner2008}, and computationally efficient algorithms have been proposed in \cite{carlsson2009topology}. These were further improved for computational speed and memory efficiency in \cite{Chen11persistenthomology} and \cite{cavanna2015}.
Motivated by this, we aim to automatically extract information on the underlying solution space.

Similar to \cite{pokorny2016}, we use filtrations of simplicial complexes to extract information about the topological structure of trajectories in a dataset---in particular, to reason about the classification of trajectories using the first homology group.
In contrast to \cite{pokorny2016}, we use Vietoris-Rips complexes for scalability and introduce filtrations based on trajectory segment distances which allows us to scale to larger datasets and further offers adaptability to particular applications through the selection of the segment distance metric.
In our experiments, we show scalability to 78-dimensional trajectories and analyze the scalability of our method.
Our method focuses on extracting equivalence classes based on the persistent topological structure of the sample data to train a memory-of-motion to initialize/warm-start optimal control planning.
We evaluate our method on initializing \gls{ddp}-style optimal control methods for cartpole and quadrotor navigation tasks. We further explore scalability through a humanoid manipulation example.
The key benefit here is to allow incorporating global information with local trajectory optimization methods \cite{pokorny2016}.

In this paper, we make the following contributions:
\begin{compactenum}
    \item We introduce a novel method for clustering time-series data based on persistent homology to identify multiple modes and discontinuities among continuous trajectories within a precomputed dataset.%
    \item We improve scalability to large datasets of time-series data by leveraging a segment-to-segment distance for filtration and use a pairwise trajectory distance for clustering. %
    \item We propose a procedure based on half-life to automatically extract the number of clusters from the persistence of cohomology groups. %
    \item We demonstrate that this method can be applied on state spaces of different dimensions and is also applicable to high-dimensional task spaces. %
    \item We evaluate the scalability of the method w.r.t. the number of trajectory samples and the number of time steps per sample.%
    \item We show that our method outperforms methods that do not exploit the clustering for warm-starting and we demonstrate this on a dynamic optimal control task in a complex environment.
\end{compactenum}

To the best of our knowledge, we are the first to describe a fully automatic trajectory clustering pipeline based on persistent homology; and the use of these tools on a dataset of motions of highly dynamic systems such as a cartpole and a quadrotor as well as to warm-start optimization solvers.

\section{Optimal Control} \label{sec:optimal_control}
We focus on discrete-time, finite-horizon nonlinear optimal control.
Consider a dynamical system for which we aim to find a policy $\vu=\pi(\vx)$ that minimizes a cost function
\begin{equation}
    J = \ell_f(\vx_T) + \sum_{t=1}^{T-1} \ell(\vx_t,\vu_t).
\end{equation}
Starting from an initial state $\vx_0$, the system evolves according to the state transition function $\vx_{t+1} = \vf(\vx_t,\vu_t)$, which incorporates the differential system dynamics and an integration scheme. $\ell_f(\vx_T)$ denotes the state cost at the end of the horizon and $\ell(\vx_t,\vu_t)$ the general running cost.
We discretize in $T$ time steps and minimize $J$ to obtain a sequence of controls $U=[\vu_1,\vu_2,\cdots,\vu_{T-1}]$, where all controls are bounded with upper and lower limits $\vu_t \in [\underline{\vu}, \overline{\vu}]$. We refer to the minimal cost for a state $\vx$ at time $t$ as the cost-to-go $V(\vx_t,T)$.

For the purpose of this work, we will solve the above optimal control problem using a direct-indirect hybridization approach \cite{mastalli2020boxfddp}, a recent extension to shooting methods with improved globalization (i.e., robustness to poor initialization) and the ability to provide initial guesses for both state and control trajectories. Because of this, we only consider general costs and bounded control constraints. We do not explicitly use constraints on the states but include these as cost terms or enforce them in the forward simulation.

\subsection{Differential Dynamic Programming}
\label{sec:ddp}
\gls{ddp} is a second-order shooting method optimizing only over the unconstrained control space displaying quadratic convergence \cite{mayne1966ddp,jacobson1968differential}. As a gradient descent method, it uses locally-quadratic approximations of the dynamics and cost functions.
\gls{ddp} alternates between a backward pass on a reference trajectory to generate a new sequence of local feedback control laws, and a forward pass computing the new state trajectory.
Following \cite{Tassa2011PhDThesis}, if $Q$ is the variation of the cost-to-go $V$ (with the subsequent cost-to-go denoted as $V'$), we can expand its variation to second-order using a Taylor series as:
\begin{equation} \label{eq:ddp_q}
Q(\delta\vx,\delta\vu)
\approx\frac{1}{2}
\begin{bmatrix}
1\\
\delta\vx\\
\delta\vu
\end{bmatrix}^\mathsf{T}
\begin{bmatrix}
0 & Q_{\vx}^\mathsf{T} & Q_{\vu}^\mathsf{T}\\
Q_{\vx} & Q_{\vx\vx} & Q_{\vx\vu}\\
Q_{\vu} & Q_{\vu\vx} & Q_{\vu\vu}
\end{bmatrix}
\begin{bmatrix}
1\\
\delta\vx\\
\delta\vu
\end{bmatrix}\text{,}
\end{equation}
where the individual terms are:
\begin{alignat}{2}
Q_{\vx} &= \ell_{\vx}+ \mathbf{f}_{\vx}^\mathsf{T} V'_{\vx} \\
Q_{\vu} &= \ell_{\vu}+ \mathbf{f}_{\vu}^\mathsf{T} V'_{\vx} \\
Q_{\vx\vx} &= \ell_{\vx\vx} + \mathbf{f}_{\vx}^\mathsf{T} V'_{\vx\vx}\mathbf{f}_{\vx}+V_{\vx}'\cdot\mathbf{f}_{\vx\vx}\\
Q_{\vu\vu} &= \ell_{\vu\vu} + \mathbf{f}_{\vu}^\mathsf{T} V'_{\vx\vx}\mathbf{f}_{\vu}+{V'_{\vx}} \cdot\mathbf{f}_{\vu \vu}\\
Q_{\vu\vx} &= \ell_{\vu\vx} + \mathbf{f}_{\vu}^\mathsf{T} V'_{\vx\vx}\mathbf{f}_{\vx} + {V'_{\vx}} \cdot \mathbf{f}_{\vu \vx}.
\end{alignat}

Solving for the optimal change in control ${\delta \vu}^*$ given a change in state $\delta \vx$ we have
\begin{align}
{\delta \vu}^* &= \argmin_{\delta \vu}Q(\delta \vx,\delta\vu) \\
&=-Q_{\vu\vu}^{-1}(Q_{\vu}+Q_{\vu\vx}\delta \vx)\text{,} \\
&= \mathbf{k} + \mathbf{K}\delta\vx
\end{align}
where $\mathbf{k}$ is the feed-forward term and $\mathbf{K}$ the feedback gain matrix.
Using this result, Equation~\eqref{eq:ddp_q} can be solved for the quadratic model of the value change:
\begin{alignat}{2}
\Delta V &= &{} -\tfrac{1}{2}Q_{\vu} Q_{\vu\vu}^{-1}Q_{\vu}\\
V_{\vx} &= Q_{\vx} & {}- Q_{\vu} Q_{\vu\vu}^{-1}Q_{\vu\vx}\\
V_{\vx\vx} &= Q_{\vx\vx} &{} - Q_{\vx\vu}Q_{\vu\vu}^{-1}Q_{\vu\vx}.
\end{alignat}

We apply regularization on the states and controls following \cite[Ch.~2]{Tassa2011PhDThesis} to ensure that the problem remains numerically well-conditioned.
To directly incorporate bound constraints on the controls without sacrificing convergence, \cite{tassa2014control} introduced the use of a projected-Newton class active-set \gls{qp} method.
Line search with variable step sizes is used to achieve convergence.
Recent improvements to \gls{ddp} introduced a hybridization called \gls{fddp} which is comparable to multiple shooting and which has shown greater globalization from poor initial guesses \cite{mastalli2020crocoddyl}.
Here, we use a recent variant that combines \cite{mastalli2020crocoddyl} and \cite{tassa2014control}, \gls{boxfddp}, to explicitly handle control bounds \cite{mastalli2020boxfddp}.

The resulting control sequence and trajectory will be locally optimal. We obtain a set of locally optimal solutions by initializing \gls{ddp} with different initial control sequences and from different initial states.

Having obtained a set of solutions to parameterized optimal control problems as a dataset, we are interested in learning a mapping $f : X \to Y$ from the problem encoding/parameterization to control trajectory output.
There is, however, no guarantee that the dataset is uni-modal nor that any of the solutions is the global optimum. Therefore, to train a warm-start model that can initialize the solver close to a global optimum, we are first interested in grouping the trajectories into uni-modal clusters.

\section{Clustering using Persistent Homology} \label{sec:persistent_homology}
Our objective is for all trajectories within a cluster to not have any topological holes (e.g., caused by obstacles) between them that would cause the warm-start to get stuck in a local minimum. In topology, such property can be described using homotopy. A homotopy between two trajectories exists if we can define a smooth continuous function that transforms one trajectory into another \cite{weisssteinHomotopy}.
Defining homotopies is an active subject of research and it is considered a difficult problem in the field of topology.
However, topological features such as clusters and holes define homology equivalences that are commonly used to approximate homotopy equivalences. 
Here, efficient algorithms exist to compute homology groups of simplicial complexes (see below). Knowledge of these homology groups allows us to reason about the global properties of a space based on local computations \cite{kaczynski2006computational}.

\begin{figure*}[ht]
    \centering
    \subfloat[Position trajectories.\label{fig:toy_example_traj}]{%
        \includegraphics[width=1.4in]{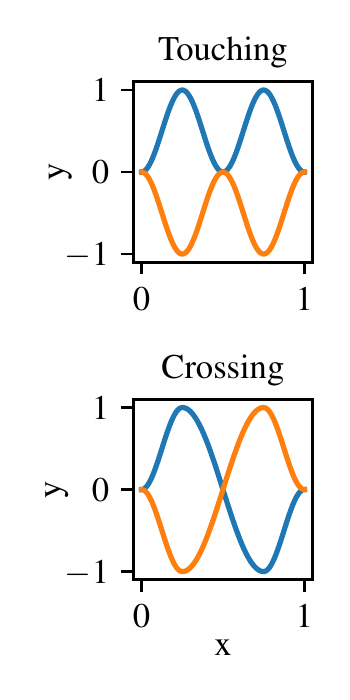}
        \vspace*{-0.075in}
    }
    \hfill
    \subfloat[Phase-space trajectories.\label{fig:toy_phase}]{%
        \includegraphics[width=1.6in]{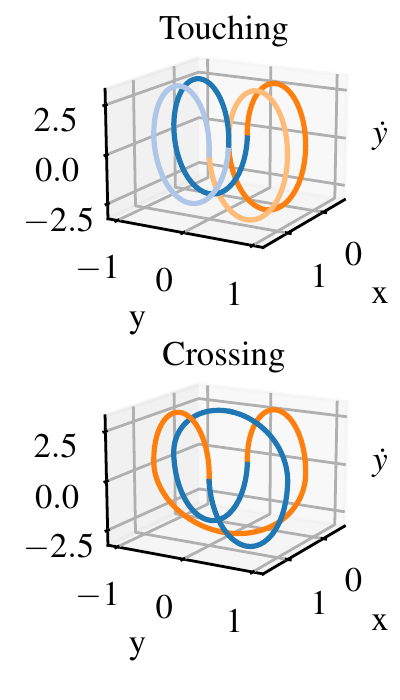}
    }
    \hfill
    \subfloat[Persistent homology.\label{fig:toy_homology}]{%
        \includegraphics[width=3.9in]{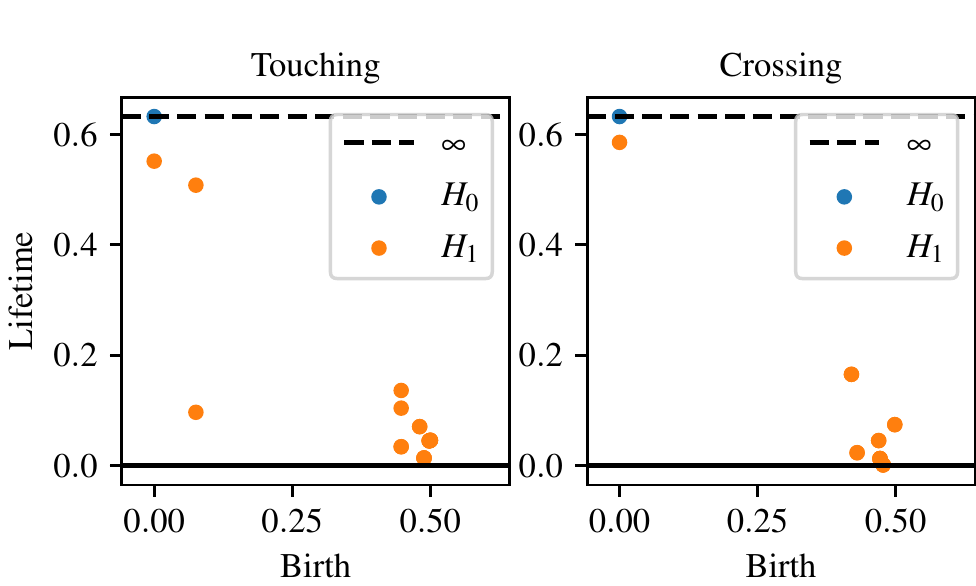}
    }
    \caption{Toy example comparing topology of a pair of trajectories (blue and orange) that touch each other at zero velocity (top) with a pair that crosses (bottom). \Cref{fig:toy_phase} shows the phase-space plot where the trajectories show a double loop and a single loop respectively. These loops are detected as two and one $H_1$ groups in the top left corner of the persistent homology plot in \Cref{fig:toy_homology}.}
    \label{fig:toy_example}
\end{figure*}

To analyze a dataset, we consider it as a set of points associated with a distance metric. The points are the state and control trajectories as computed by an optimal control method, discretized in time.
From these points, we are particularly interested in computing the invariant \emph{cohomology} of the output space. This can be interpreted as the number of ``holes'' that would separate the data into clusters.
A key building block of algebraic topology are \emph{simplicial complexes} made up of simplices: Points are $0$-simplices, edges/lines between two points form a $1$-simplex, three points forming a triangular face are a $2$-simplex, and so on for higher-dimensional simplices.
During filtration, we increase the distance (or scale) parameter $r$. For each point in the dataset, we connect all points within distance $r$ to form simplices.
We record when simplices emerge, e.g. two points ($0$-simplices) get connected by an edge, forming a $1$-simplex. This event is also called the \emph{birth} of a simplex. At the same time, one of the $0$-simplices disappears as it merges into the larger connected component (the edge). This is marked as a \emph{death} of the shorter-lived simplex. We plot these events on the persistence diagram by a dot along the \emph{Birth} and \emph{Death} axis. The events are color-coded by the dimension of the simplex or complex that died in the event ($0$-simplices - $H_0$, $1$-simplices - $H_1$, etc.) As the value of $r$ increases, we track how simplices get merged into larger complexes and the persistence of the data is revealed by the distance of the death event from the birth (along the vertical axis).
Alternatively, we can plot a persistence diagram as in \Cref{fig:toy_homology} where the $x$-axis denotes the \emph{Birth} and the $y$-axis shows the \emph{Lifetime} ($\mathrm{Lifetime}=\mathrm{Death}-\mathrm{Birth}$).
Invariant features of the underlying dataset have a long persistence/lifetime (are higher up) and can thus be read off the homology diagram.
Alternatively, these features can be visualized using a barcode diagram \cite{ghrist2008barcodes}.
Here, the rank of the zeroth-dimensional homology group ($H_0$) corresponds to the number of connected components, while the rank of the first-dimensional homology group ($H_1$, the number of one-dimensional ``holes'') allows us to reason about the number of clusters in the data.

\subsection{Dealing with time-series data} \label{sec:homology_time_series_data}
As the trajectories represent time-series data (the state or controls at time steps $t_1, \ldots, t_T$), we cannot directly use them as a dataset for filtration.

Suppose we have $N$ samples of $M$-dimensional time-series data with $T$ time steps.
As we are interested in the structure of the $M$-dimensional space, a na\"ive point-cloud approach would be to stack the samples into an $NT$-dimensional vector to create a pairwise distance matrix of dimension $(NT)\times(NT)$.
However, we have to choose a large number of samples and time steps to cover the solution space sufficiently densely. One way to reduce the size of the distance matrix without sacrificing the coverage is to compute the distance between trajectory segments instead of the trajectory knot points. We propose to use linear segments and calculate the distance as proposed in \cite{LUMELSKY198555}. This allows us to approximate the distance matrix of a set of trajectories with relatively few time steps which decreases the size of the distance matrix to $(N(T-1))\times(N(T-1))$.
While this change does not immediately reduce the size of the distance matrix significantly, it allows us to adaptively replace several data points that are approximately linear with a single line segment (i.e., adaptive numbers of $T$).

After we obtain a full distance matrix, we post-process it by incorporating explicit connectivity information from the trajectories:
We explicitly set the distance for subsequent time steps to zero as suggested by \cite{pokorny2016}. We can do the same for connected start and end states.
This further ensures that all samples are connected into one $H_0$ group and works similarly to the common start and end points proposed by \cite{pokorny2016}. As a result, we always get a single $H_0$ component in all our persistence plots.
We then apply filtration to the post-processed distance matrix to extract the persistent homology groups.

To illustrate, we apply this process to a toy problem, see \Cref{fig:toy_example}.
We generated two pairs of trajectories using sine waves. Each trajectory is a sequence of 2D positions and corresponding 2D velocities, resulting in a trajectory in $\mathbb{R}^4$. \Cref{fig:toy_example_traj} shows the position trajectory. 
The first pair of trajectories are touching in the middle, where both the position and velocity equal to zero. The second pair is crossing in the middle with the position being zero but the velocities being large. 
We have designed the trajectory such that the velocity along the $x$-axis is constant. This allows us to visualize the 4D phase-space plot by dropping the velocity along the $x$-axis and plotting the velocity along the $y$-axis as the $z$-coordinate in the 3D plot in \Cref{fig:toy_phase}. This plot shows how the trajectories that are touching create two loops, while the trajectories that are crossing create only one loop in the phase-space.
The persistent homology plot in \Cref{fig:toy_homology} shows this topology. On the left, we see two $H_1$ groups (orange dots), and on the right, one $H_1$ group that corresponds to the number of loops in the phase-space of each pair of trajectories. There are also several short-lived $H_1$ groups close to the bottom of the graph. These are small holes that correspond to the aliasing artifacts. They depend on the resolution at which we sampled the sine waves and can be considered as noise. In both cases, there is one $H_0$ group (blue dot) that captures the single connected component created by connecting the ends of the two trajectories.

We are interested in $H_1$ groups with early birth (furthest to the left on the diagram) and long lifespan (higher up). These translate into persistent features in the data that we want to preserve.
We can additionally extract the separating distance as $>0.3$ (above where the short-lived groups disappear but below the long-lived groups appear).

Note, however, that the proposed process considers distance matrices in their dense form, which can very quickly exhaust memory during computation.%
\footnote{We have explored the use of filtration using sparse distance matrices \cite{cavanna2015}, however, the temporal structure of time-series data is not compatible with this technique.}
In practice, we frequently reduce the sampling frequency along the time dimension to make the problems tractable. The toy example trajectory can be sampled at as few as $T=5$ time steps without changing the topology of the dataset.

An alternative approach would be to use dimensionality reduction tools or alternate representation with embeddings before filtration.
Appropriate alternate representations are crucial as they can significantly change the topology of the space. This is especially crucial when dealing with time-series data that includes state derivatives and dimensions with various units. This is common with robot joint position and velocities having different scales, and with linear joints, angular joints, and the floating base being measured in different units. We manually set the relative scaling and use the persistent homology to analyze the topology of the resulting space. In the toy problem, this corresponds to scaling the $z$-axis in \Cref{fig:toy_phase}. Once we have selected the scaling, we proceed to use the same tools for clustering the dataset.
We explore this concept further using alternate space representations in \Cref{sec:evaluation_different_filtration_spaces}.

\subsection{Extraction of number of clusters from persistence of cohomology groups} \label{sec:homology_persistence_extraction}
The lifetime (persistency) of simplicial complexes is commonly visualized in a persistence diagram as in \Cref{fig:toy_homology}. To automatically extract the number of significant separating radii, we propose to use a heuristic based on half-life between subsequent lifetimes in an ordered list.
We show this procedure in \Cref{alg:extract_number_of_classes_from_h1}. Two parameters are the cut-off ratio between subsequent lifetimes (we use $0.8$) and the minimum lifetime (distance) to filter out short-lived simplicial complexes (we use $0.1$).

We ignore the $H_0$ group since all trajectories always belong to a single connected component due to how we treat the time-series data.
We count the number of $H_1$ groups and compute the number of clusters as $H_1 + 1$. We use this number as an input to the clustering algorithm.

\begin{algorithm}[ht]
    \caption{Algorithm for extracting the number of clusters based on the first homology group.}
    \label{alg:extract_number_of_classes_from_h1}
    \begin{algorithmic}[1]
    \renewcommand{\algorithmicrequire}{\textbf{Input:}}
    \renewcommand{\algorithmicensure}{\textbf{Output:}}
    \REQUIRE $H_1$ groups; $\mathrm{cutoff\_ratio}$; $\mathrm{min\_lifetime}$
    \ENSURE  $\mathrm{num\_classes}$
     \STATE $\mathrm{num\_classes = 1}$
     \STATE $\mathrm{lifetimes = SortDescending(Deaths-Births)}$
     \STATE $\mathrm{previous\_lifetime = infinity}$
     \FOR {$\mathrm{lifetime} \in \mathrm{lifetimes}$}
        \IF {($\mathrm{lifetime} < \mathrm{min\_lifetime} \textbf{ or } \newline \mathrm{lifetime} < \mathrm{cutoff\_ratio}*\mathrm{previous\_lifetime}$)}
            \RETURN $\mathrm{num\_classes}$
        \ELSE
            \STATE $\mathrm{num\_classes} += 1$
            \STATE $\mathrm{previous\_lifetime} = \mathrm{lifetime}$
        \ENDIF
     \ENDFOR
    \RETURN $\mathrm{num\_classes}$
    \end{algorithmic}
\end{algorithm}

\subsection{Clustering} \label{sec:homology_clustering}
Knowing the number of classes in the dataset, we now need to apply labels to each trajectory sample. To achieve this, we create a trajectory-wise distance matrix of size $D_\textrm{pairwise-trajectories} \in \mathbb{R}^{N\times N}$.
To create it, we apply the process using the trajectory segment distance and post-processing for time-series data described in \Cref{sec:homology_time_series_data} for each pair of trajectories $i,j$ to obtain a pairwise post-processed distance matrix $D_{ij}$. This is the same distance matrix we use for computing the persistent homology. We then extract the maximum of the minimum distances across trajectory segments $D_{ij}$ and store it as the representative distance between trajectories $i$ and $j$ in $D_\textrm{pairwise-trajectories}$.
This metric has a topological meaning. Since we already enforce the distance between subsequent time steps and the start and end segments to be zero, the lifetime of a $H_1$ group is proportional to the furthest distance between two trajectories computed using our proxy metric. The trajectory-wise distance matrix, therefore, serves as an approximation of a metric based on the persistent homology of a pair of trajectories. As a result, it encodes how close the two trajectories are to being in the same homotopy class.

Finally, using the number of clusters from \Cref{alg:extract_number_of_classes_from_h1} and with the precomputed distance matrix $D_\textrm{pairwise-trajectories}$, we use agglomerative clustering with the single linkage method to assign class labels to the original dataset.

\section{Mixture of Experts} \label{sec:moe}
Whether, and how quickly, the optimization solver converges from a given initial condition and parameterized goal setting depends significantly on the quality of the initial guess.
If we can provide a good initial guess to the optimization solver, considerable speed-up for the convergence can be achieved \cite{merkt2018leveraging}.
Having separated the data into clusters with continuous input-output relationships, we now propose a Mixture-of-Experts approach to predict the state and control trajectories given new initial conditions.%
\footnote{Traditionally, shooting methods can only be initialized using control trajectories. Using \gls{fddp}-derived solvers, we can provide both state and control trajectories as an initial guess.}
We propose to use $k$ experts, where $k$ is the number of clusters identified in \Cref{sec:homology_persistence_extraction}, and a gating network to decide which expert to query. Note, that multiple experts can also be explored independently and simultaneously, similar to ensemble methods.
Here, we use simple \gls{mlp} models with ReLu activation for each expert and train using Adam \cite{kingma2015adam}. We did not tune the hyper-parameters apart from ensuring that the cumulative capacity of each of the compared methods matches.

\section{Evaluation}
\label{sec:experiments}
We test our methodology on optimal control tasks using the cartpole and quadrotor dynamic models.
We implement the optimal control problem, system dynamics, and \gls{boxfddp} solver in the \gls{exotica} \cite{exotica}.
For topological analysis, we compute the persistence cohomology of the dataset using the efficient Ripser library \cite{bauer2017ripser,ctralie2018ripser}.
All evaluations were carried out on a laptop using a single core of an Intel Core i7-9850H CPU at \SI{4.2}{\giga\hertz} and \SI{64}{\giga\byte} \SI{2933}{\mega\hertz} memory.

We provide our implementation, datasets, and notebooks as open-source for reproducing our results.%
\footnote{\url{https://github.com/wxmerkt/topological_memory_clustering}}

\subsection{Scalability} \label{sec:evaluation_scalability}
\begin{figure}[t]
    \centering
    \includegraphics[width=\linewidth,trim={0.41cm 0.55cm 0.42cm 0.39cm},clip]{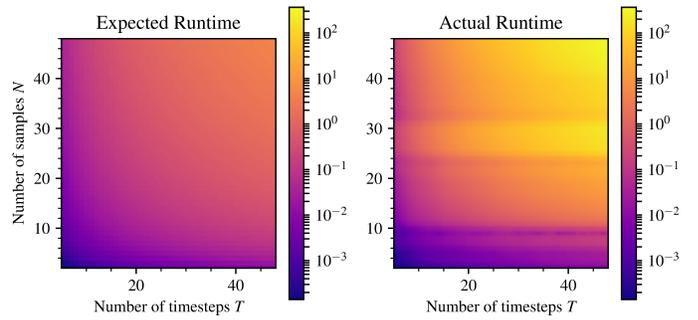}
    \caption{\emph{Left:} Expected computation time (\si{\second}) scaling with $\mathcal{O}( (N\,(T-1))^2)$. \emph{Right:} Computation times from filtration of datasets with varying $N$ and $T$. We use the cartpole dataset from \Cref{sec:evaluation_cartpole_swingup} and cubic interpolation for time re-sampling.}
    \label{fig:cartpole_scalability}
\end{figure}

First, we look at the scalability of our proposed method to identify the number of clusters (\Cref{sec:homology_time_series_data}) with i) the number of samples in the dataset ($N$), ii) the length of the trajectories ($T$), and iii) the dimension of the state space.
The size of the distance matrix used for filtration scales with $N$ and $T$. Hence, we expect the computational complexity to scale with $\mathcal{O}( (N\,(T-1))^2)$. We use the cartpole dataset from \Cref{sec:evaluation_cartpole_swingup} and cubic interpolation for time re-sampling to evaluate scalability with $N$ and $T$. The results are presented in \Cref{fig:cartpole_scalability} and show filtration time scaling as expected.
Note, the use of line segment distances enables us to achieve scalability to larger datasets by replacing densely sampled trajectory sub-segments with a single segment-wise distance (e.g., a piecewise linear approximation).

\begin{figure}[b]
    \subfloat[Visualization of the two modes for the swing-up task.\label{fig:cartpole_two_modes}]{
        \includegraphics[width=.495\linewidth,trim={5cm 11cm 5cm 3cm},clip]{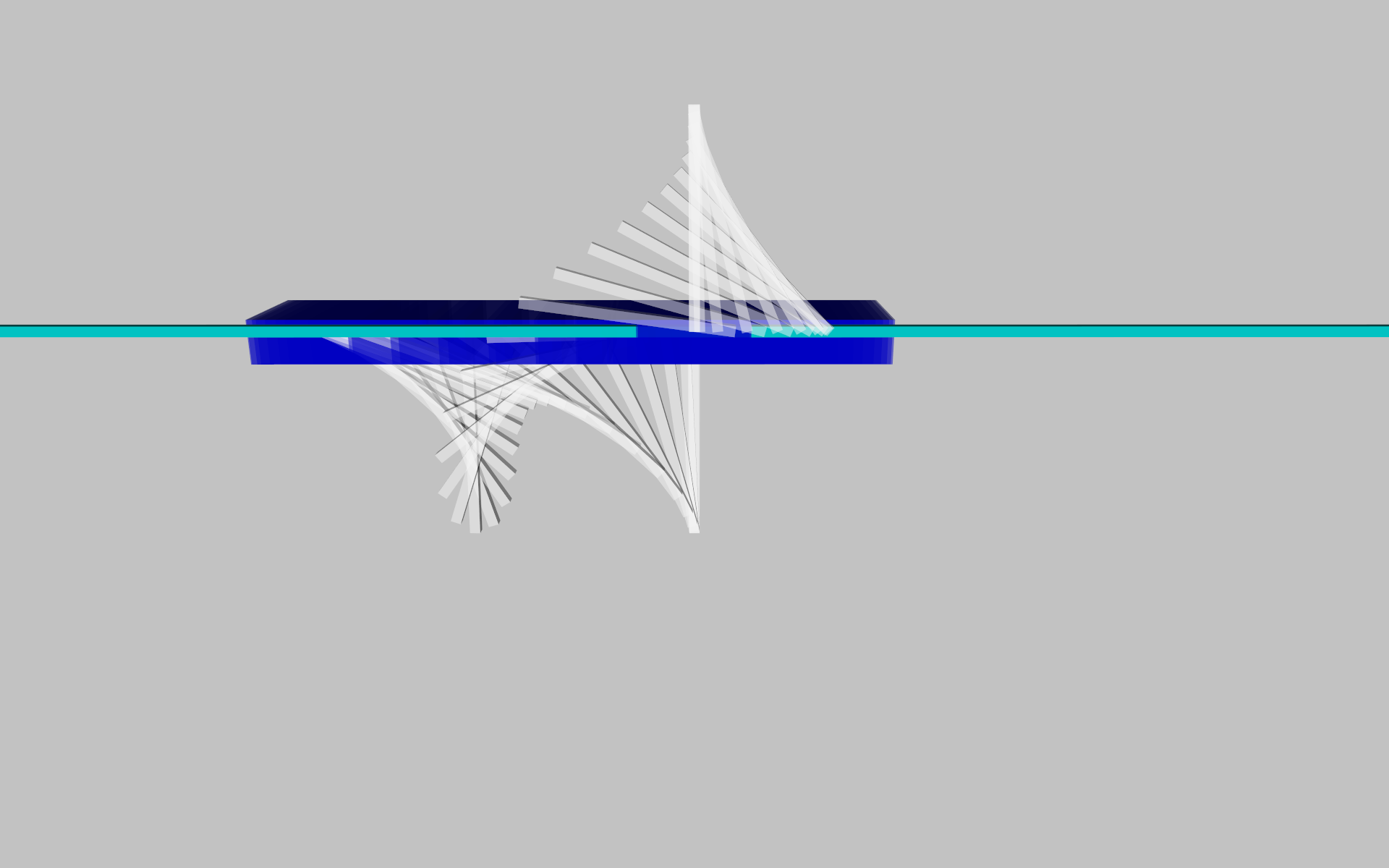}
        \hfill
        \includegraphics[width=.495\linewidth,trim={5cm 11cm 5cm 3cm},clip]{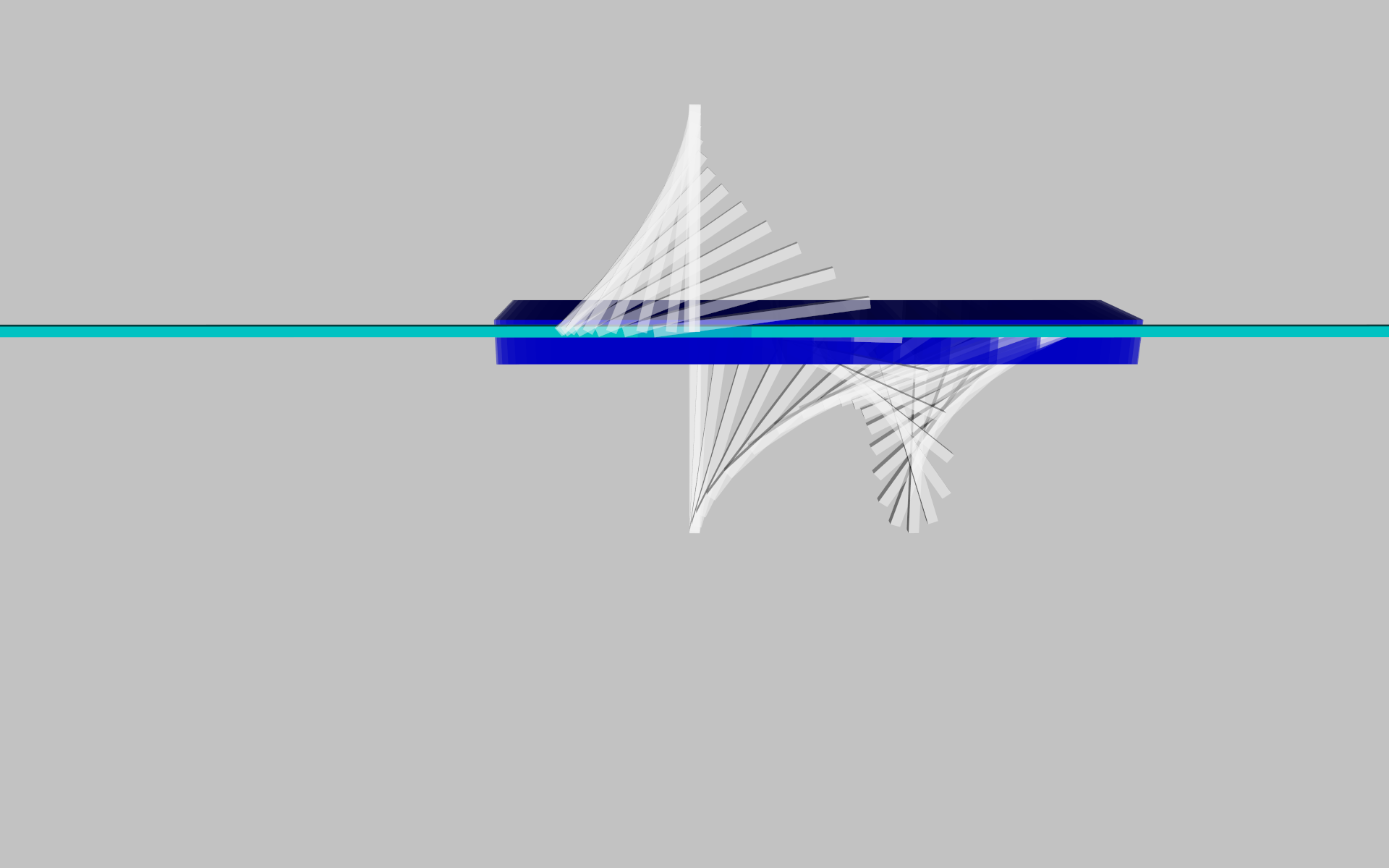}
    }
    \vfill
    \subfloat[%
    State trajectories for swing-up policies from random initial states.
    \label{fig:cartpole_state_trajectories_raw}]{
        \includegraphics[width=\columnwidth]{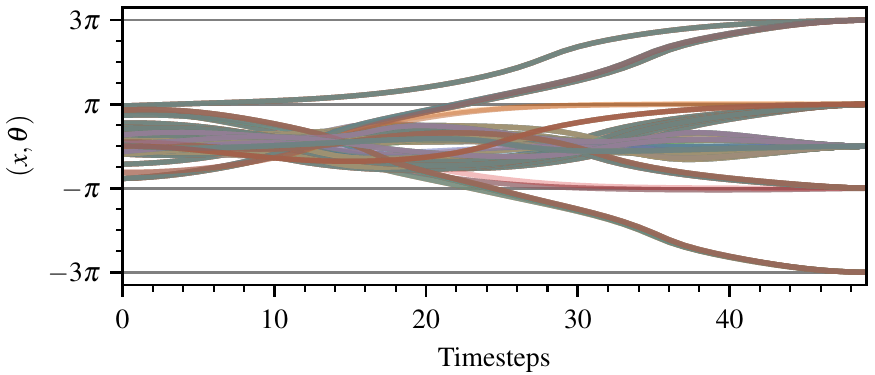}
    }
    \caption{The cartpole swing-up task: We illustrate the two modes (\Cref{fig:cartpole_two_modes}) and show 500 policies for a cartpole swing-up task from random initial states (\Cref{fig:cartpole_state_trajectories_raw}). Note, depending on the initial velocity the pole may complete several rotations before stabilizing.}
    \label{fig:cartpole_swingup}
\end{figure}

\subsection{Cartpole swing-up} \label{sec:evaluation_cartpole_swingup}
The cartpole is a dynamic system where a pole is mounted with an unactuated hinge joint on a cart that travels on a rail. It uses horizontal forces as controls $\vu$.
Due to control limits and as an under-actuated system, the cartpole is a canonical task for nonlinear optimal control as the cart needs to gather energy in order to be able to swing up.
We model the system following \cite{tedrake2014underactuated} with the slider position denoted as $x$, the angle of the pole as $\theta$, and the state as $[x,\theta,\dot{x},\dot{\theta}]^\mathsf{T}$. We limit the control input to $\vu \in [-10,10]$ \si{\newton}.
The aim is to swing up the pole to the upright position with the cart at the origin and zero final velocity (i.e., $\vx_{goal}=[0,\pm\pi,0,0]^\mathsf{T}$).
Note, that we do not require the pole to reach the final value of $\pi$, but any configuration that is upright: It is irrelevant from which side the pole swings up---or whether it completes more than one full rotation prior to coming to rest.
Therefore, we model the continuous hinge joint using the special orthogonal group $\mathbb{SO}(2)$ and represent the state as $[x, \cos\theta,\sin\theta,\dot{x},\dot{\theta}]^\mathsf{T}$.

We now consider a swing-up task on the cartpole. From intuition, we can postulate that there are two modes: \emph{swinging up from the left side} and \emph{swinging up from the right side}. Indeed, these two solutions can be seen in \Cref{fig:cartpole_two_modes}.

To build the dataset, we randomly sample start states in the range $\pm(\SI{1}{\metre},\SI{\pi}{\radian},\SI{1}{\metre\per\second},\SI{0.5\pi}{\radian\per\second})$ and control policies uniformly from $[-1,1]$ \si{\newton}. We solve the optimal control problem using \gls{boxfddp} until we have $N=10$ solutions per random start state. 
The state trajectories in \Cref{fig:cartpole_state_trajectories_raw} show that the pole may complete several full rotations prior to reaching a zero velocity and getting into the upright pose.

\begin{figure*}[t]
    \centering
    \subfloat[Persistent homology filtration for the cart-pole swing-up dataset: Distance matrix and persistent homology diagram.\label{fig:cartpole_persistent_homology}]{%
        \includegraphics[width=.49\linewidth]{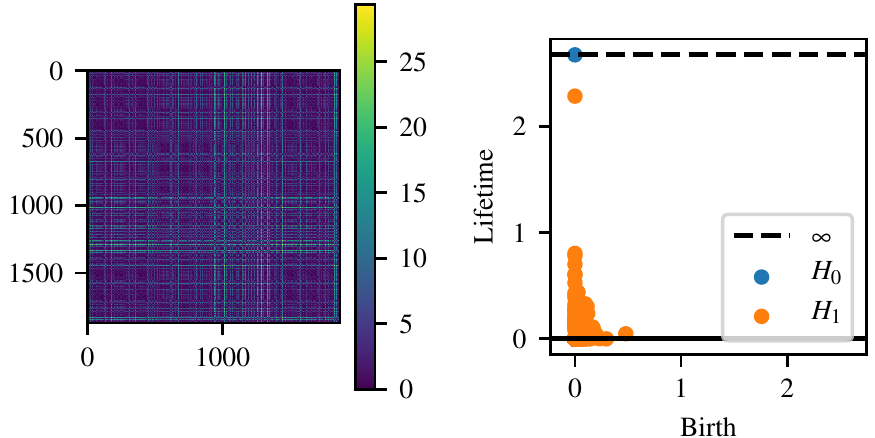}
    }
    \hfill
    \subfloat[Cart-pole state trajectories labeled using persistent homology: Random start states are shown with red circles and the swing-up goal in green.\label{fig:cartpole_state_trajectories_labelled}]{%
        \includegraphics[width=.49\linewidth]{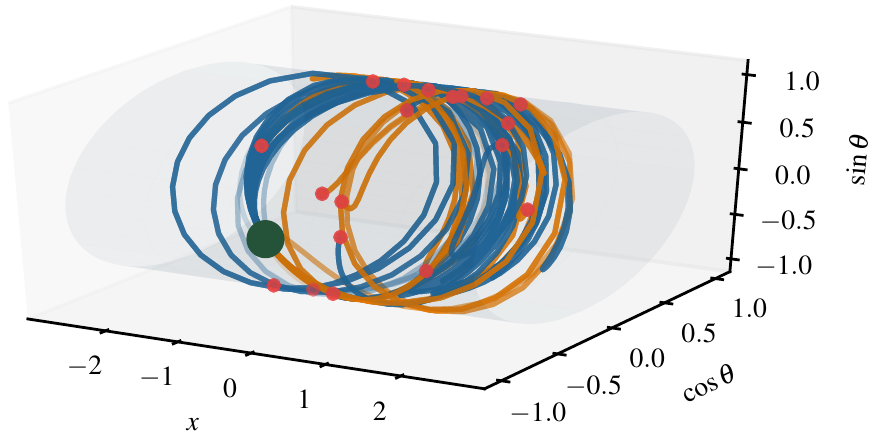}
    }
    \\\vspace{12pt}
    \subfloat[Control trajectories for random start states with class labels assigned from the output of the persistent homology filtration.\label{fig:cartpole_control_trajectories_labelled}]{%
        \includegraphics[height=1.5in,trim={0in 0in 3.35in 0in},clip]{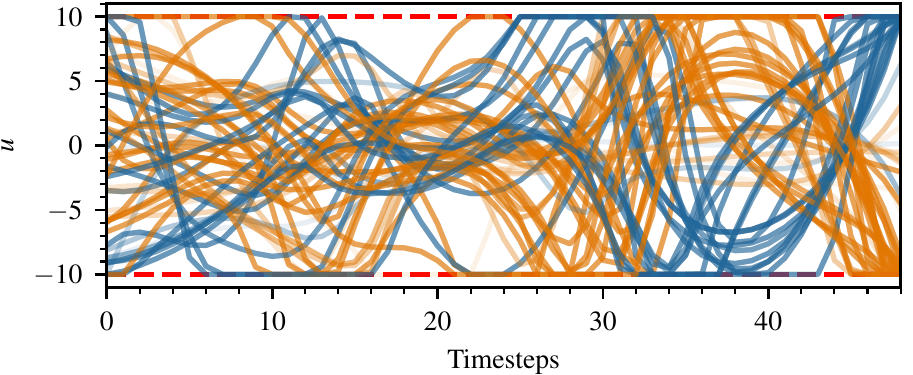}
    }
    \hfill
    \subfloat[Mean and standard deviation of the control trajectories for each of the classes (orange, blue) and the dataset (black).\label{fig:cartpole_control_policy_means}]{%
        \includegraphics[height=1.5in,trim={3.65in 0in 0in 0in},clip]{figures/cartpole/cartpole_class_labels_and_aware_mean.pdf}
    }
    \caption{Results on the cart-pole swing-up task: (a) Persistent homology diagram computed from the pre-processed distance matrix showing one hole (two classes) as indicated by a single $H_1$ group group (orange dot in the left top corner). (b) State trajectories labeled using persistent homology. (c) Control trajectories labeled using persistent homology. (d) Mean and standard deviation of control trajectories conditioned by class highlighting that without class-sensitivity, the information cancels out.
    }
    \label{fig:cartpole_results}
\end{figure*}

We now aim to extract (a) the number of topological classes contained in the dataset and (b) assign the corresponding class labels.
From the obtained persistent homology diagram in \Cref{fig:cartpole_persistent_homology}, we see that (a) there is one connected set and (b) that there is one hole, i.e., that there are two classes. We can further read off a separating distance of $\approx 1.5$.

Using this information, it is now possible to cluster the raw state and control trajectory data, cf. Figures \ref{fig:cartpole_state_trajectories_labelled} and \ref{fig:cartpole_control_trajectories_labelled}.

We investigate how separating the solutions into unimodal clusters impacts the warm-start strategies based on simplistic interpolation, and strategies relying on training generative models from the raw control trajectory samples. We now look at the overall mean and standard deviation of the control trajectories and compare this to the mean of each of the two identified and labeled classes (cf. \Cref{fig:cartpole_control_policy_means}). 
As shown in the figure, if the multi-modality of the solutions is not taken into account, the information in the data effectively cancels out producing a mean trajectory (black) that encodes little useful structure and has a large standard deviation.
This is due to the symmetry in the control trajectories---by summing them all terms vanish.

\subsection{Quadrotor navigation in a maze} \label{sec:evaluation_quadrotor_warm-starts}
In the previous experiment, we tested our methodology on a low-dimensional problem using a cartpole without considering environments with obstacles, where collision avoidance is needed. 
We now consider a set of experiments to investigate both the scalability in terms of problem dimensionality as well as to include multi-modality in the solutions introduced by collision avoidance.
Additionally, we show the impact of the data separation on successfully learning initial seeds for the optimal control problem.

Quadrotors are agile multi-rotor drones that have become very popular in the past decade due to increased battery energy densities, more powerful motors, and reduced component prices.
To model the quadrotor dynamics, we follow \cite{mellinger2012trajectory} with minor modifications: 
We do not consider the effects of air drag (or near ground effects) and control the rotor forces directly. Hence, the control inputs are $\vu \in \mathbb{R}^4$ (limited between \SIrange{0}{5}{\newton}) and the state is $\vx \in \mathbb{R}^{12}$ using Euler representation for the angular component of the floating base.

We sample a dataset of trajectories for a quadrotor flying from a start state uniformly sampled from $[-3.25,-0.25]$ for each $x$, $y$, and $z$ to a goal position ($(x,y,z)=(1.75, 1.75, 1.75)$) while avoiding collision in an environment using the formulation described in \cite{merkt2018leveraging}, see \Cref{fig:discontinuity_multimodality}.
The environment consists of three intersecting cylinders centered at the origin.
We use $T=50$ knot points with $\Delta t=0.05\si{\second}$.
We initialize the optimal control solver with a state trajectory obtained from RRT-Connect~\cite{kuffner2000rrtconnect} and a control trajectory for hovering with noise sampled from the standard normal distribution ($\mathcal{N}(0,0.01)$) and solve until convergence.

\begin{figure*}[t]
    \centering
    \includegraphics[width=.8\linewidth]{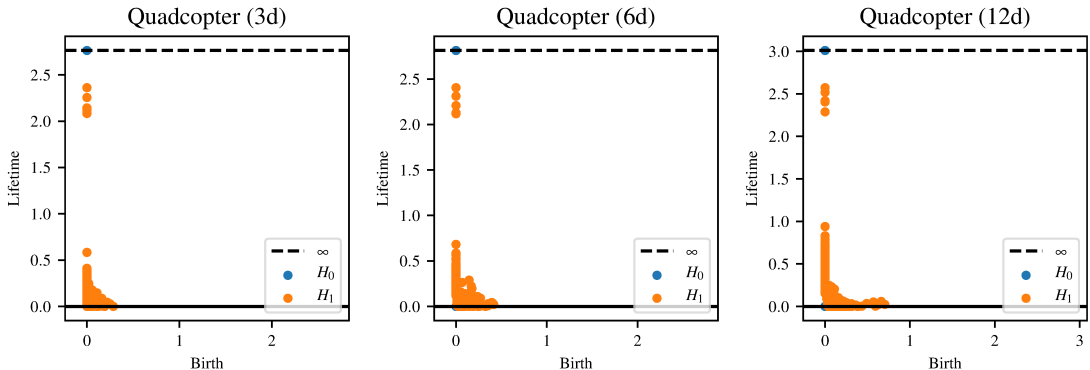}
    \caption{Filtration of the quadrotor dataset using different state spaces: translation only, translation and rotation, and full state space including velocities: The identified underlying topology and computation times are the same.}
    \label{fig:homology_quadrotor_dense_filtration}
\end{figure*}

Our dataset consists of $5535$ valid state and control trajectories.
We depict the persistent homology graphs in \Cref{fig:homology_quadrotor_dense_filtration} and automatically extract the number of clusters as six (five holes).
Filtration on 3D (base position), 6D (base position and orientation), and 12D (base position, orientation, and spatial velocities) state spaces took \SI{18.9}{\second}, \SI{18.5}{\second}, and \SI{22.0}{\second}, respectively.
Analogously to the cartpole example, we perform clustering using persistent homology. The labeled state trajectories are shown in \Cref{fig:discontinuity_multimodality}.

\begin{figure*}[t]
    \centering
    \includegraphics[width=\linewidth,trim={0.3cm 0.3cm 0.3cm 0.25cm},clip]{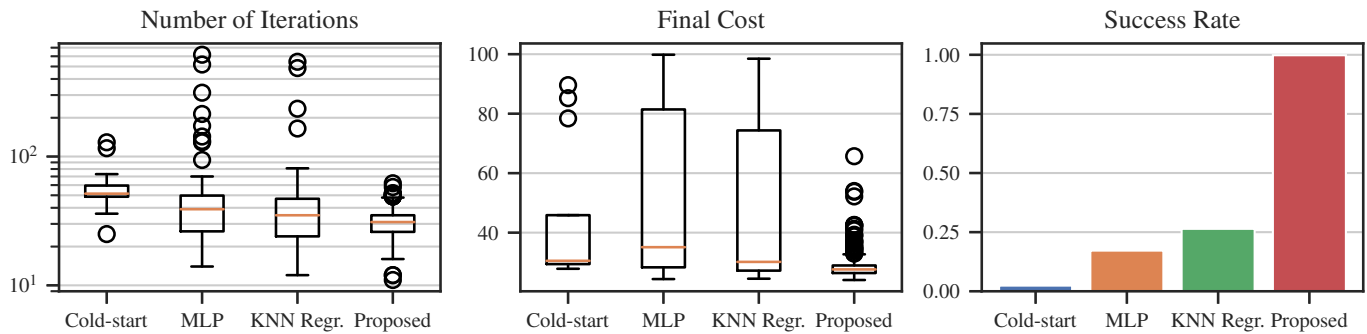}
    \caption{Warm-start comparison results on the quadrotor: Number of major iterations, final cost, and convergence success rate when initializing the optimal control problem using different initialization methods. We compute the number of iterations and final cost based on successful trials only.}
    \label{fig:quadrotor_results}
\end{figure*}

\begin{figure}[htb]
    \includegraphics[width=\linewidth,trim={0.35cm 0.38cm 0.34cm 0.34cm},clip]{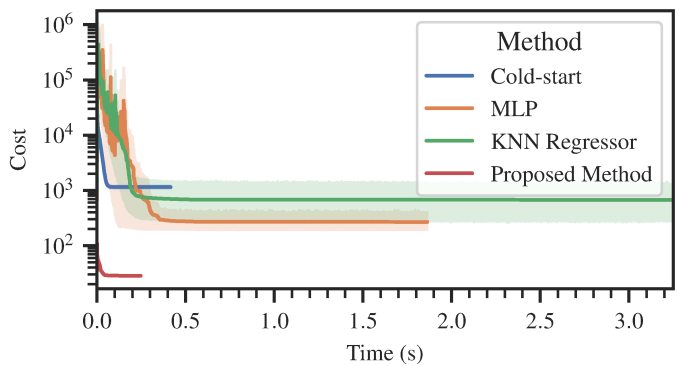}
    \vspace{-1.5em}
    \caption{Cost evolution (mean, standard deviation) for solving the quadrotor task using \gls{boxfddp} and state and control trajectory initializations predicted by different learning methods.}
    \label{fig:quadrotor_benchmark_cost_vs_time}
\end{figure}

We now use the labeled samples to learn state and control predictors to generate initial seeds for our optimal control solver following \Cref{sec:moe}.
To benchmark, we randomly sample start states from the same range as in the dataset and predict state and control trajectories using the four comparison methods:
\begin{compactenum}
    \item Cold-start: Controls initialized to achieve hovering and states initialized with the initial state for entire horizon.
    \item \gls{mlp}: Prediction from a single \gls{mlp} each for both state and control trajectories (1-hidden layer, 200 neurons, $245,596$ trainable parameters).
    \item KNN-Regressor: Prediction from a KNN-Regressor method which averages over $k$ neighbors. $k \in [1,10]$ has been brute-forced for best performance using a held-out validation set.
    \item Our method: A Mixture-of-Experts setup trained on the separated data. Smaller \gls{mlp} where trained for each of the continuous subsets and a gating network (with softmax activation) trained to select the most suitable expert (1-hidden layer, 50 neurons). The total trainable parameters across all experts and gating network are $252,472$. Note, this is the only class-aware prediction method in the test field.
\end{compactenum}
We perform this benchmark for $500$ random start states and show the mean and standard deviation of the convergence (cost evolution versus wall-clock time) in \Cref{fig:quadrotor_benchmark_cost_vs_time}.
It is evident that the modality-aware method has a lower initial cost and faster convergence than other initialization methods. Note that the cold-start (no initial guess) performs better than a non-mode-aware learning method (\gls{mlp}, KNN-Regressor).
A likely explanation for this is that a roll-out of the predicted control trajectories has large dynamic defects compared with the predicted state trajectories.
This can be seen by comparing the learned state initialization, the roll-out of the learned control initialization, and the final optimized trajectory in the accompanying video.%
\footnote{\url{https://youtu.be/lUULTWCFxY8}}
We have also evaluated the number of major iterations, final cost, and success rate as shown in \Cref{fig:quadrotor_results}. The number of major iterations show the same trend as the convergence time in \Cref{fig:quadrotor_benchmark_cost_vs_time}, which is expected. The success rates for the four methods (cold-start, MLP, KNN regressor, and proposed) were $2.4\%$, $17.2\%$, $26.4\%$, and $99.8\%$, respectively.

\subsection{Humanoid Reaching: Filtration in alternate state spaces} \label{sec:evaluation_different_filtration_spaces}
In addition to clustering initial seeds for dynamical optimal control, persistent homology can also be used as a tool to choose an alternate task space in which to represent the trajectory. We have illustrated this problem using the toy example in \Cref{sec:homology_time_series_data} where we talked about scaling of the velocity component, which is a trivial way of defining an alternate space. In more general scenarios, we look for spaces where the motion would produce simpler and more persistent topology. This is particularly relevant when we intend to track a kinematic trajectory using a Proportional-Derivative (PD) controller. A PD controller minimizes error in joint space. It is however very common to implement an operational space controller which is a PD controller minimizing error in task space. To ensure the controller can keep tracking the task, we want to choose a space in which the PD controller can track a kinematic reference trajectory by minimizing a Euclidean error. A space with a simpler topology would therefore be more suitable for tracking a reference trajectory.

\begin{figure}[!htb]
    \includegraphics[width=.495\linewidth,trim={14cm 3.7cm 14cm 2.6cm},clip]{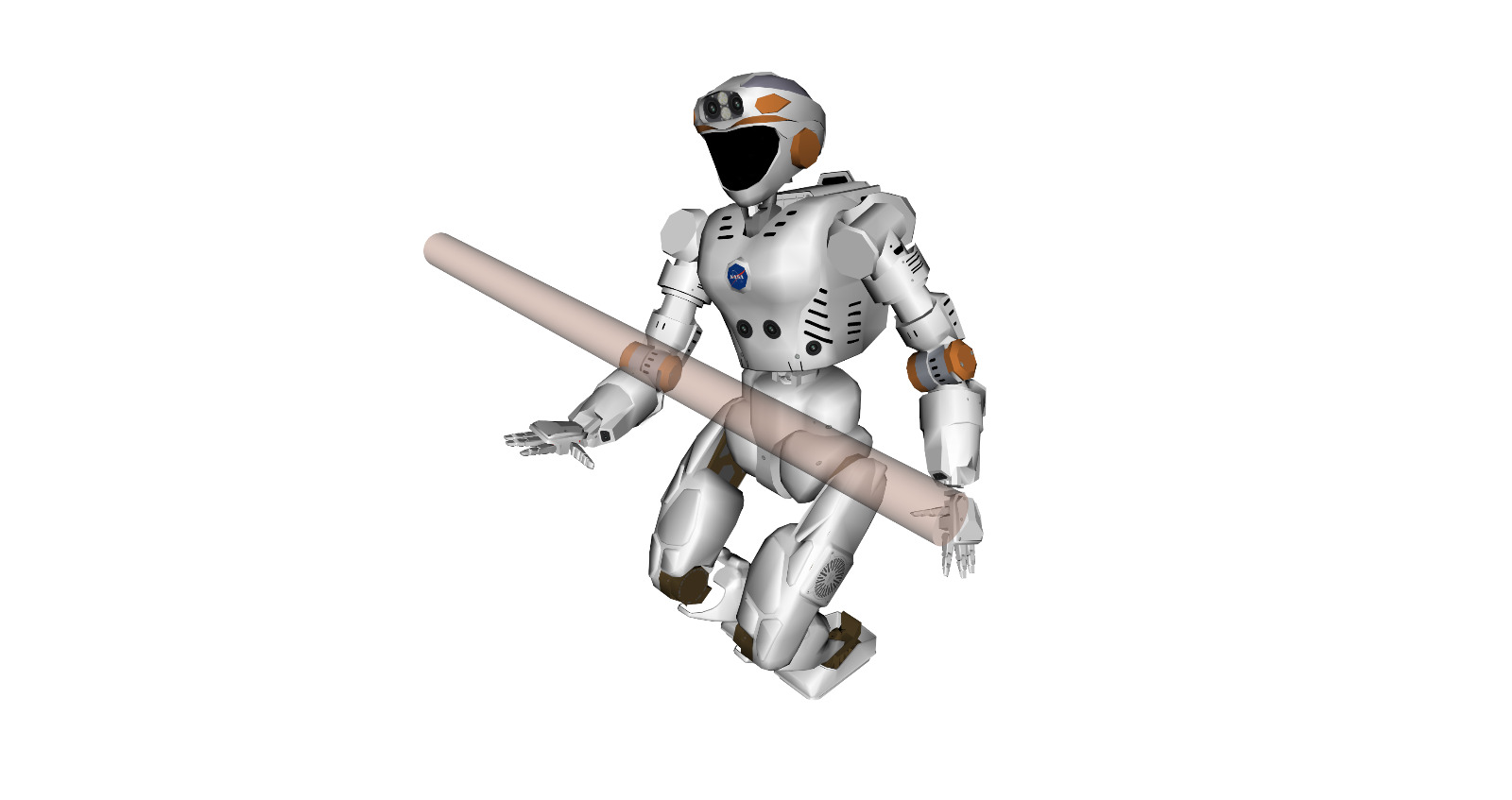}
    \hfill
    \includegraphics[width=.495\linewidth,trim={14cm 3.7cm 14cm 2.6cm},clip]{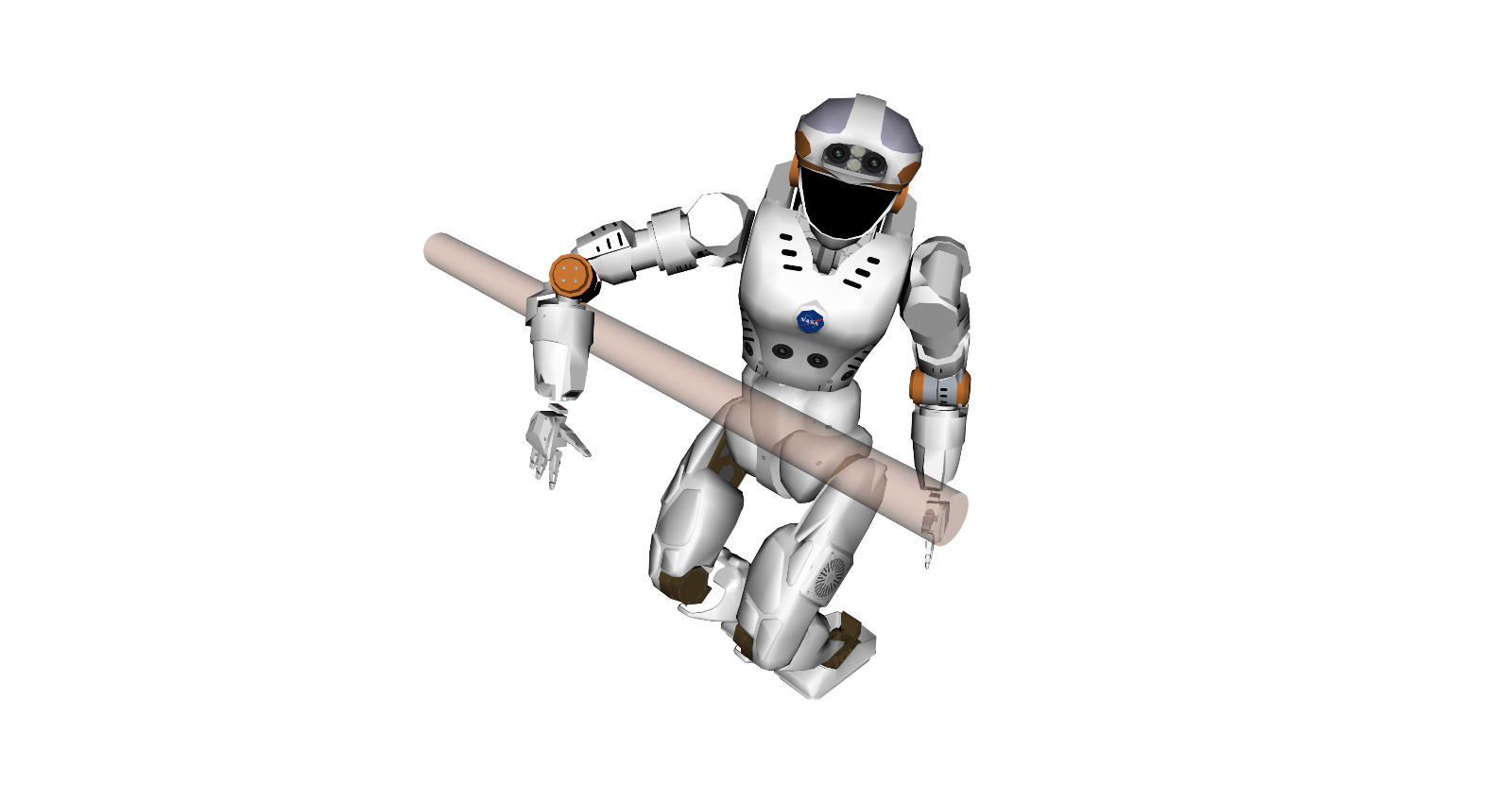}
    \caption{Valkyrie robot reaching scenario, showing two distinct reaching poses creating the bifurcation in the solution space.}
    \label{fig:val-reaching}
\end{figure}

We have generated a dataset of kinematic reference trajectories for the torso and arm of the Valkyrie humanoid robot (see \Cref{fig:val-reaching}). We have planned 100 trajectories sampled at 200 knot points per trajectory. We have defined a reaching task with the reaching target in front of the robot but behind a horizontal bar. The task has two solutions: reaching above and below the bar. We have solved the problem using RRT-Connect to ensure that the trajectories are collision-free.

\begin{figure}[!htb]
    \includegraphics[width=\linewidth,trim={0.62cm 0.4cm 0.72cm 0.35cm},clip]{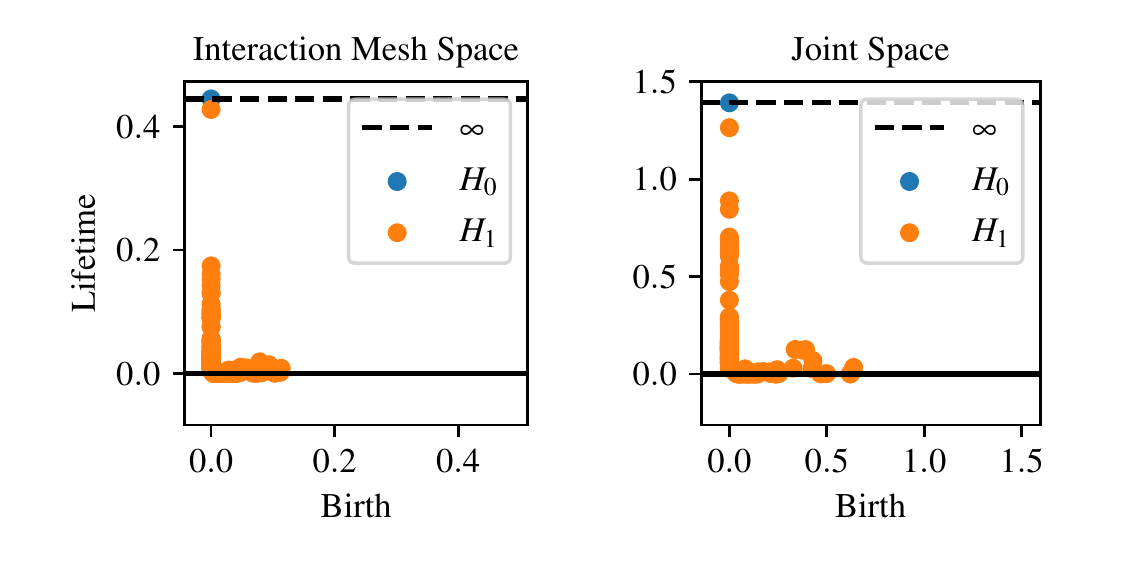}
    \vspace{-0.5em}
    \caption{Persistence intervals for the interaction mesh space (left) and the joint space (right).}
    \label{fig:persistence-imesh}
\end{figure}

\Cref{fig:persistence-imesh} shows the results of filtration in joint space and an alternate, topological state space representation: the interaction mesh \cite{ivan2013topology}. The interaction mesh is hereby defined as the edges between points on the robot and key points in the environment capturing the relationship between the movement of the robot within its surroundings by computing the Laplace coordinates of these points. The coordinates in the interaction mesh space also have a much more uniform relative scale along each dimension which means that the persistent topology features are a lot more likely to emerge at the same scale across the whole space.
The results show that the interaction mesh space representation has a much more clearly defined topology than the joint space. In the persistence diagram, this is visible as one $H_1$ group far away from a cluster of short-lived noise. As a result, the topology of the trajectory data is more clearly represented in interaction mesh space, highlighting an interesting avenue for further work exploring the topology of other task spaces.

\section{Conclusions}
In this paper, we introduced a method to automatically identify the number of classes and cluster a dataset of trajectories using persistent homology.
We then trained a generative model to compress and generalize the dataset for use as an initial seed in future optimizations.
Our experiments explored this concept on optimal control problems involving dynamics as well as high-dimensional kinematic tasks and focused on establishing whether relevant topological information can be extracted to assist the encoding and learning stages of storing a memory-of-motion.

Our results confirm that considering the underlying topological features of the dataset is important and that tools from algebraic homology can be used to guide clustering.
Indeed, exploring multiple modes can be important for warm-starting (cf. \Cref{fig:quadrotor_benchmark_cost_vs_time}) and further allows the development of ensemble methods that explore multiple warm-start guesses in parallel \cite{lembono2020memory}.
While these simultaneously explore multiple warm-starts, an alternative approach would be to test ranked warm-start modes/hypotheses akin to sequence prediction \cite{dey2013contextual}.

\subsection{Limitations and Future work}
One limitation of our approach is that the tools for filtration are very sensitive to the amount of data. Empirically, for a current laptop computer, distance matrices up to a few thousand rows/columns take tens of seconds to compute the persistent homology.
However, it is important to note that, in practice, the time dimension can often be sub-sampled while preserving the salient features in the data.
A key consideration here is that the cohomology computation can only capture features that are present in the data: If a topological hole is not captured in the discretized (or subsampled) data, a clustering method will not be able to extract it. A common example is thin obstacles for which trajectories may be collision-free at the discretized knot points but would require traversing through the obstacle to connect subsequent trajectory points.
We have previously studied and proposed a solution to this challenge in \cite{merkt2019continuous}.
Since our method proposes the use of a segment-to-segment distance, it permits the evaluation of distances in either discrete- or continuous-time allowing for these considerations to be incorporated.
Additionally, the underlying structure of the dataset (i.e., number of classes) can be explored using subsets of the data without affecting the scalability of the clustering step---a feature we have explored and applied to scale to larger datasets. 
In this work, we did not explore homology groups in dimensions greater than 1. The $H_2$ groups, in particular, can be used to cluster trajectories enclosing volumes of space, such as a sphere. We did not explore such motion here as it would require a large number of trajectories to produce a dense enough coverage for the $H_2$ persistence to be reliably detected. This is an interesting future research direction, especially for systems and tasks with redundancy and periodicity.

The algorithms we used in this work for computing the homology groups did not take advantage of parallelism and available implementations are memory-intensive.
Recent advances in computational homology have focused on leveraging massively parallel architectures to reduce the computation time by one to two orders of magnitude \cite{zhang2020ripserplusplus,morozov2020towards}. We plan to evaluate and leverage these in future work.

One way to address dataset size could be to apply curve fitting techniques to reduce the dimension of the distance matrix prior to filtration. Curve fitting techniques and embeddings (e.g., splines or Bezier curves) can be used to represent longer time-series segments in place of the linear segments we have used in \Cref{sec:homology_time_series_data}.
Alternatively, \cite{Perea2015} previously applied sliding windows to discover periodicity in time-series data using persistent homology.
It is worthwhile to explore whether a similar approach can be applied in our case instead of the dense filtration of the full dataset.

If the trajectories cover the solution space sparsely, the topology of the data may depend more on the random seed of the trajectory generator rather than the underlying solution space. This means that in some cases, a relatively small obstacle may be mistaken for a non-persistent $H_1$ group and result in clustering the data incorrectly. This could lead to a warm-start trajectory passing through an obstacle. We could remedy this by either increasing the number of trajectories in the dataset, by utilizing a collision avoidance technique, or by artificially inflating the distance between a pair of trajectories passing around the obstacle. However, the former comes with increased computational cost while the latter is only possible if we know that the obstacle is present. Enforcing topology around known obstacles would be an interesting extension of this work.

State spaces including derived quantities such as velocities or accelerations pose a further challenge. To normalize/trade-off spaces with different units (i.e., scaling velocity down w.r.t. configuration), we apply intuition and manually scale each dimension. Latest work on multi-parameter persistent homology \cite{multi-parameter-homology} offers new tools to compute the persistence of the trajectories with different amounts of scaling in a principled way. This problem could then be further expanded to analyzing the topology of arbitrary parameterized task spaces to discover spaces with a simpler topology that is favorable for simple operational space controllers.

Another challenge, also identified by others, is that we did not consider changing environments. Some authors \cite{mansard2017irepa,tang2019discontinuity} outlined the possibility to enhance the problem parameterization with information on the size and location of geometric primitive shaped objects. We consider this inflexible, as fixed-size, basic representations always limit expressiveness and require vast numbers of samples to explore the increased dimensionality of the problem space. 
Instead, \cite{jetchev2013fast} investigated and compared learned situation descriptors. More recent work focused on learning latent space representations directly from sensor data such as point clouds \cite{qureshi2021motion}.
Along with large, labeled datasets on 3D objects and shapes, these are promising avenues for further investigation.

Finally, we explored systems with continuous dynamics and straight-forward start and goal situations. It would be interesting to explore the use of tools from algebraic topology on tasks with discontinuous dynamics and periodicity, for instance, locomotion on legged platforms.

\bibliographystyle{IEEEtran}
\bibliography{IEEEfull,IEEEconf_original,bib}

\begin{IEEEbiography}[{\includegraphics[width=1in,height=1.25in,clip,keepaspectratio]{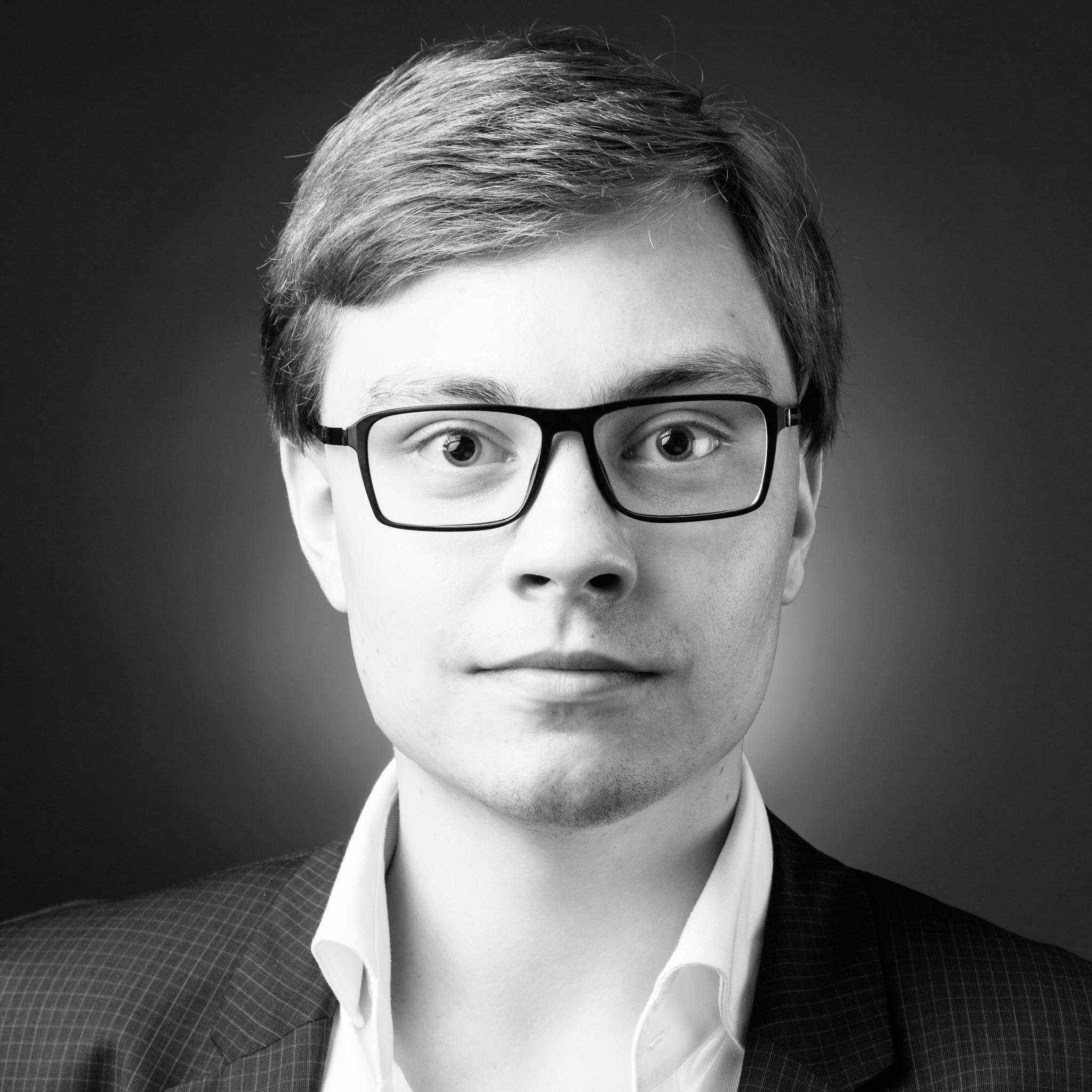}}]{Wolfgang Merkt}
received the B.Eng.(Hns) degree in mechanical engineering with management and the M.Sc.(R) and Ph.D. degrees in robotics and autonomous systems from the University of Edinburgh in 2014, 2015, and 2019, respectively. He is currently a postdoctoral researcher at the Oxford Robotics Institute, University of Oxford. During his Ph.D., he worked on trajectory optimization and warm-starting optimal control for high-dimensional systems and humanoid robots. Wolfgang's research interests include fast optimization-based methods for planning and control, loco-manipulation, and legged robots.
\end{IEEEbiography}

\begin{IEEEbiography}[{\includegraphics[width=1in,height=1.25in,clip,keepaspectratio]{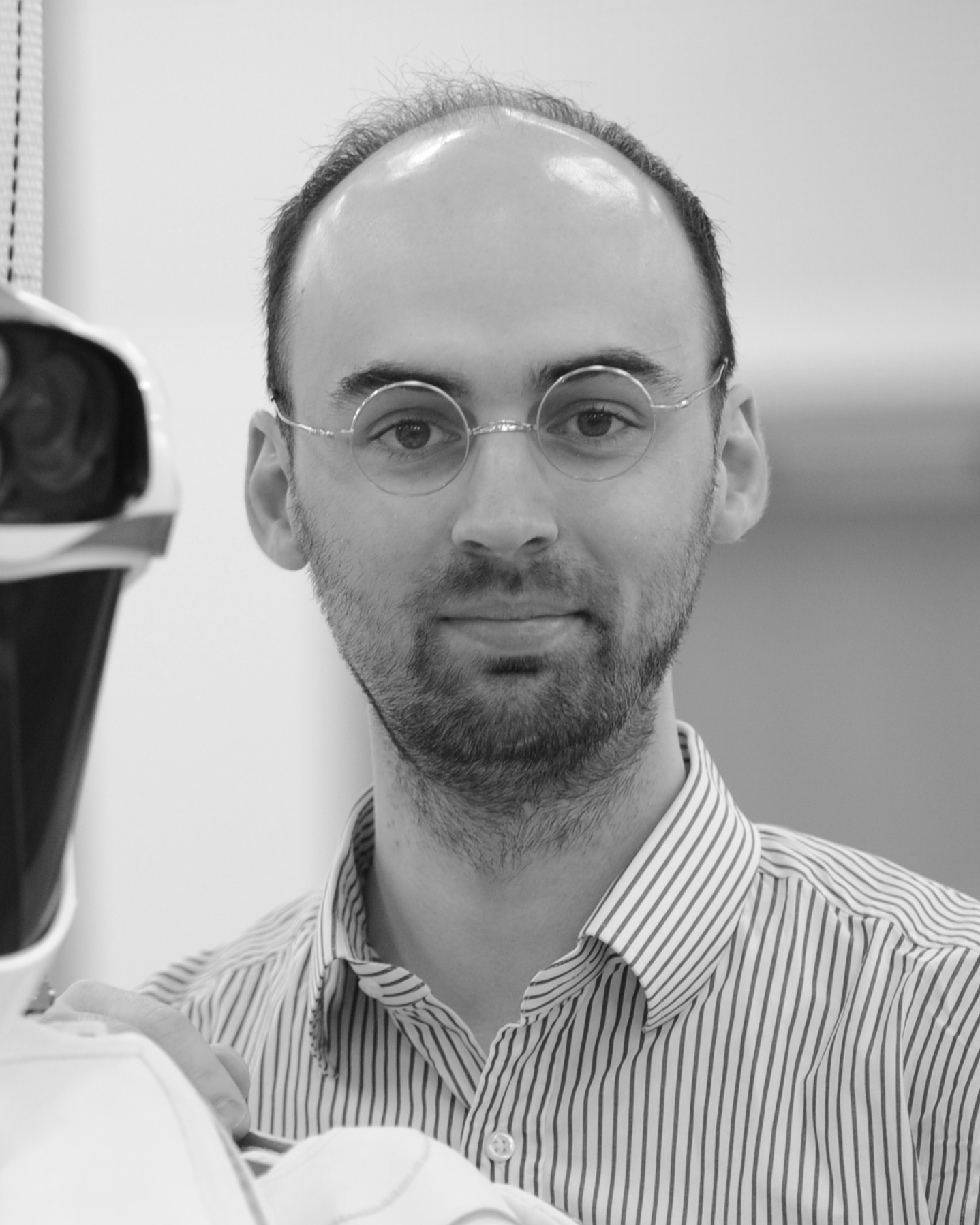}}]{Vladimir Ivan}
received his Ph.D. on the topic of Motion synthesis in topology-based representations at the University of Edinburgh where he is currently working as a Senior Researcher in the School of Informatics. He has previously received a M.Sc. in Artificial Intelligence specializing in Intelligent Robotics at the University of Edinburgh and a B.Sc. in AI and Robotics from the University of Bedfordshire. Vladimir's research interests are motion planning and modeling, topology, humanoid and legged robotics, space robotics, and shared autonomy.
\end{IEEEbiography}

\begin{IEEEbiography}[{\includegraphics[width=1in,height=1.25in,clip,keepaspectratio]{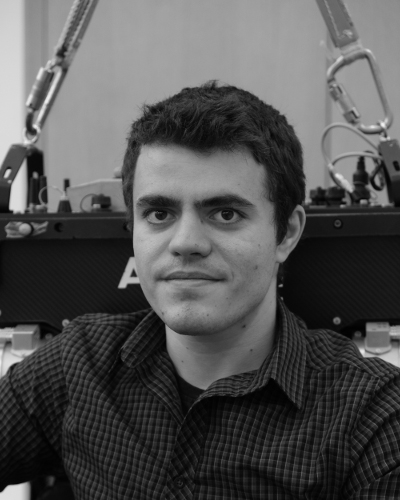}}]{Traiko Dinev}
 received a B.Sc. degree in computer science (with electronics) and an M.Sc. in robotics and autonomous systems from the University of Edinburgh. He is currently pursuing a PhD degree with the Edinburgh Centre for Robotics at the University of Edinburgh. Traiko's research interests include motion planning for dynamic locomotion, hardware-software co-design for planning and control, and applied statistical machine learning. 
\end{IEEEbiography}

\begin{IEEEbiography}[{\includegraphics[width=1in,height=1.25in,clip,keepaspectratio]{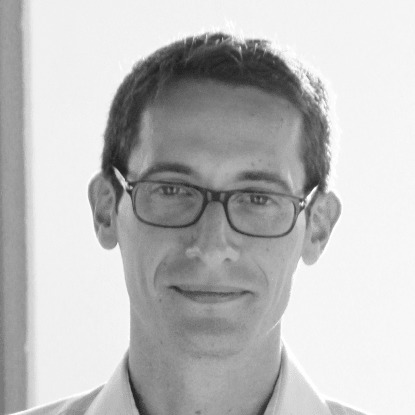}}]{Ioannis Havoutis}
is a Lecturer in Robotics at the University of Oxford. He is part of the Oxford Robotics Institute and a co-lead of the Dynamic Robot Systems group. His focus is on approaches for dynamic whole-body motion planning and control for legged robots in challenging domains. From 2015 to 2017, he was a postdoc at the Robot Learning and Interaction Group at the Idiap Research Institute. Previously, from 2011 to 2015, he was a senior postdoc at the Dynamic Legged System lab at the Istituto Italiano di Tecnologia. He holds a Ph.D. and M.Sc. from the University of Edinburgh.
\end{IEEEbiography}

\begin{IEEEbiography}[{\includegraphics[width=1in,height=1.25in,clip,keepaspectratio]{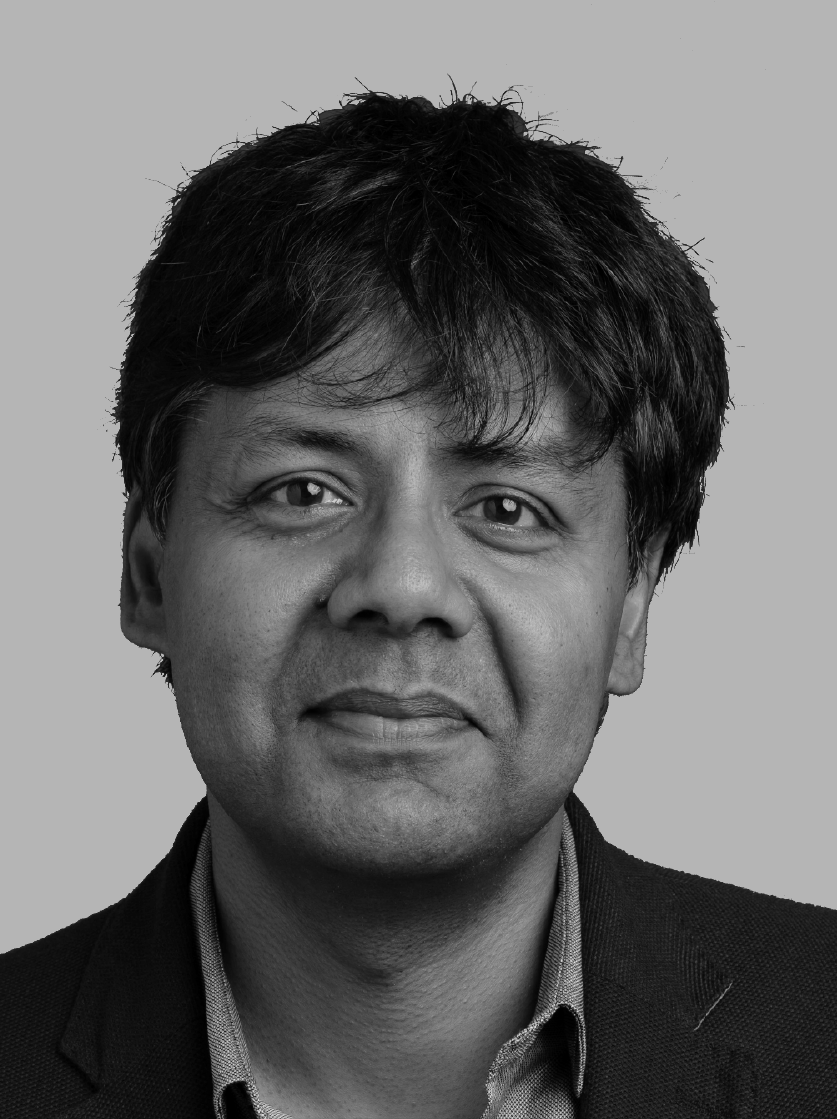}}]{Sethu Vijayakumar}
is the Professor of Robotics at The University of Edinburgh, U.K., an Adjunct Faculty with the University of Southern California, Los Angeles, CA, USA and the Founding Director of the Edinburgh Centre for Robotics. His research interests include statistical machine learning, anthropomorphic robotics, planning, multi-objective optimization and optimal control in autonomous systems as well as the study of human motor control. Prof. Vijayakumar helps shape and drive the national Robotics and Autonomous Systems (RAS) agenda in his role as the Programme co-Director for Artificial Intelligence (AI) with The Alan Turing Institute, the UK's national institute for data science and AI. He is a Fellow of the Royal Society of Edinburgh, a Judge on BBC Robot Wars and winner of the 2015 Tam Dalyell Prize for excellence in engaging the public with science.
\end{IEEEbiography}

\end{document}